%% file: main.tex
\documentclass[nohyperref]{article}
\usepackage{microtype}
\usepackage{graphicx}
\usepackage{subfigure}
\usepackage{booktabs} % for professional tables
\usepackage[utf8]{inputenc}
\usepackage{amsfonts}
\usepackage{standalone}
\usepackage{bm}
\usepackage{comment}
\usepackage{amsmath}
\usepackage{natbib}
\usepackage{tikz}
\usetikzlibrary{bayesnet}
\usepackage{xcolor}
\usepackage[ruled,vlined ]{algorithm2e}
\usepackage[title,toc,titletoc,page]{appendix}
\def\*#1{\boldsymbol{\bm{#1}}}

\newcommand{\lin}{\text{lin}}
\newcommand{\nonlin}{\not{\lin}}
\newcommand{\beginsupplement}{%
        \setcounter{table}{0}
        \renewcommand{\thetable}{S\arabic{table}}%
        \setcounter{figure}{0}
        \renewcommand{\thefigure}{S\arabic{figure}}%
     }
\usepackage{hyperref}
\usepackage[accepted]{icml2022}
\usepackage{amsmath}
\usepackage{amssymb}
\usepackage{mathtools}
\usepackage{amsthm}
\usepackage{authblk}

\usepackage[capitalize,noabbrev]{cleveref}

\theoremstyle{plain}

\theoremstyle{definition}

\theoremstyle{remark}

\usepackage[textsize=tiny]{todonotes}

\icmltitlerunning{Modelling Technical and Biological Effects in scRNA-seq data with Scalable GPLVMs}

\begin{document}

% \twocolumn[
\onecolumn
\icmltitle{Modelling Technical and Biological Effects in single-cell RNA-seq data with Scalable Gaussian Process Latent Variable Models (GPLVMs)}

\icmlsetsymbol{equal}{*}

\begin{icmlauthorlist}
\icmlauthor{Vidhi Lalchand}{equal,phy}
\icmlauthor{Aditya Ravuri}{equal,comp}
\icmlauthor{Emma Dann}{equal,sanger}
\icmlauthor{Natsuhiko Kumasaka}{sanger}
\icmlauthor{Dinithi Sumanaweera}{sanger}
\icmlauthor{Rik G.H. Lindeboom}{sanger}
\icmlauthor{Shaista Madad}{sanger}
\icmlauthor{Sarah A. Teichmann}{sanger,cav}
\icmlauthor{Neil D. Lawrence}{comp}

% %\icmlauthor{}{sch}
% %\icmlauthor{}{sch}
\end{icmlauthorlist}

\icmlaffiliation{sanger}{Wellcome Sanger Institute; Wellcome Genome Campus, Cambridge, UK}
\icmlaffiliation{cav}{Theory of Condensed Matter, Cavendish Laboratory, Department of Physics, University of Cambridge; Cambridge, UK.}
\icmlaffiliation{comp}{Department of Computer Science, University of Cambridge; Cambridge, UK}
\icmlaffiliation{phy}{Department of Physics, University of Cambridge, Cambridge, UK}

\icmlcorrespondingauthor{Vidhi Lalchand}{vr308@cam.ac.uk}
% % \icmlcorrespondingauthor{Firstname2 Lastname2}{first2.last2@www.uk}

% % You may provide any keywords that you
% % find helpful for describing your paper; these are used to populate
% % the "keywords" metadata in the PDF but will not be shown in the document
% \icmlkeywords{gaussian processes, latent variable models, computational biology, single-cell RNA}

% \vskip 0.3in
% ]

% this must go after the closing bracket ] following \twocolumn[ ...

% This command actually creates the footnote in the first column
% listing the affiliations and the copyright notice.
% The command takes one argument, which is text to display at the start of the footnote.
% The \icmlEqualContribution command is standard text for equal contribution.
% Remove it (just {}) if you do not need this facility.

%\printAffiliationsAndNotice{}  % leave blank if no need to mention equal contribution
% \printAffiliationsAndNotice{\icmlEqualContribution} % otherwise use the standard text.
\printAffiliationsAndNotice{\icmlEqualContribution}

\begin{abstract}
% The Gaussian process latent variable model \citep{lawrence2004gaussian} is a popular approach for probabilistic non-linear dimensionality reduction and has been used frequently in analysis of single-cell data \citep{ahmed2019grandprix, campbell2015bayesian, verma2020robust} and pseudotime estimation. In this work we leverage stochastic variational inference (SVI) \citep{JMLR:v14:hoffman13a} for mini-batch learning within the GPLVM framework \citep{lalchand2022generalised} and further extend the model to explicitly account for donor variation and other confounders like technical batch effects. \ed{Emma/Dinithi to update} We demonstrate strong results on a single-cell dataset of size 22K cells (rows) and 5K genes (columns/dimensions) used in \citep{kumasaka2021mapping} where we recreate the latent space visualisation observed in \citep{kumasaka2021mapping} at a fraction of the training time (20x speed-up.) (to add perhaps - something along the lines of) We also explore interpretability of latent variables (enforced due to kernel design and initialization) in (COVID data + severity, etc.)

Single-cell RNA-seq datasets are growing in size and complexity, enabling the study of cellular composition changes in various biological/clinical contexts. 
Scalable dimensionality reduction techniques are in need to disentangle biological variation in them, while accounting for technical and biological confounders. In this work, we extend a popular approach for probabilistic non-linear dimensionality reduction, the Gaussian process latent variable model \citep{lawrence2004gaussian}, to scale to massive single-cell datasets while explicitly accounting for technical and biological confounders. The key idea is to use an augmented kernel which preserves the factorisability of the lower bound allowing for fast stochastic variational inference.
We demonstrate its ability to reconstruct latent signatures of innate immunity recovered in \citet{kumasaka2021mapping} with 9x speed-up on training time.
We further analyse a dataset of blood cells from COVID-19 patients and demonstrate  
that this framework enables to capture interpretable signatures of infection, while integrating data across individuals and technical batches.
Specifically, we explore COVID-19 severity as a latent dimension to refine patient stratification and capture disease-specific gene expression signatures.
\end{abstract}

\section{Introduction}

The development of single-cell transcriptomics technologies (scRNA-seq) has led to the discovery of new cell types in tissues and of cellular composition changes associated with disease, aging or experimental perturbations. Dimensionality reduction and unsupervised learning techniques are widely adopted to identify patterns in scRNA-seq data, capturing complex biological phenomena (e.g. cell differentiation). With datasets growing in size and complexity (often including many samples generated across multiple laboratories, conditions and individuals), models that resolve both  technical and biological confounders (e.g. batch effect, inter-individual variation,  proliferation signatures) are needed to learn interpretable representations of cell states.

The Gaussian Process Latent Variable Model (GPLVM) has been previously used as a non-linear probabilistic method for dimensionality reduction, trajectory inference and noise reduction in  single-cell data (\citet{campbell2015bayesian, buettnerComputationalAnalysisCelltocell2015, ahmed2019grandprix, verma2020robust}). This model allows for uncertainty in inferences and incorporation of additional information on biological processes through priors, for example to model spatio-temporal dependencies between cells (\citet{svenssonSpatialDEIdentificationSpatially2018a, veltenIdentifyingTemporalSpatial2022}). However, existing implementations either lack the ability to account for confounding covariates or are too inefficient for application to large datasets.

In this work, we extend the GPLVM to attain  scalability across large scRNA-seq datasets through an augmented kernel function which jointly accounts for both technical and biological covariates, and successfully demonstrate its ability to capture biologically interpretable signatures from an innate immunity dataset \cite{kumasaka2021mapping} and a COVID infection dataset \cite{stephensonSinglecellMultiomicsAnalysis2021}. 

\section{Methods}

\subsection{Model} % https://www.overleaf.com/project/605e07104b6c6221cfa7a557

The GPLVM uses a Gaussian process (GP) to define a non-parametric mapping from a set of latent variables to gene expression. The covariance function of the GP controls the inductive biases of the mapping, controlling properties like smoothness and periodicity. %(A review of Gaussian processes is presented in \citet{rasmussen2004gaussian}). 
(See \citet{rasmussen2004gaussian} for an extensive review on GPs). 
Let the gene expression data
%\footnote{A log one plus transformation is applied to the count data.}
be represented by $\{\bm{y}_{n}\}_{n=1}^{N} \equiv Y \in \mathbb{R}^{N \times D}$ where $N$ denotes the number of cells and $D$ the number of genes. We wish to capture the most significant factors of variation in $Y$ %this data set 
by learning a $Q$ dimensional latent encoding per cell, $\{\bm{x}_{n}\}_{n=1}^{N} \equiv X \in \mathbb{R}^{N \times Q}$ where $Q < D$ provides dimensionality reduction.
%We use an additive model, with random effects explaining the variation associated with technical or biological confounders and a Bayesian GPLVM \citep{titsias2010bayesian} to model latent effects.

The core component of this model %such models 
is a sparse GP \cite{hensman2013gaussian} that models latent effects, %and it is possible to add 
 also allowing to include 
additional terms for random effects associated with %a known set of
known technical or biological confounding covariates %(represented as a 
(via a design matrix $\Phi$) as in \citet{kumasaka2021mapping},
%The data likelihood resembles a random effects model with the core component modelled by a matrix variate sparse GP and additional terms which capture the random effects across genes ($D$) and cells ($N$) 
\begin{equation}
    Y = \underbrace{F}_{\textrm{\tiny{{Sparse GP}}}} + \underbrace{\Phi}_{\textrm{\tiny{design matrix}}}\times\underbrace{B}_{\textrm{\tiny{random effects}}} + \underbrace{\bm{\epsilon}}_{\textrm{\tiny{noise model}}}. \hspace{2mm}
    \label{nats}
\end{equation}
The model above renders the likelihood non-factorisable (see section \ref{nonfactor}), preventing truly scalable inference through mini-batching. In the sections below, we interpret the additive model above as a GP with an augmented kernel function that jointly models the process $F + \Phi B$. Then, replacing this joint GP with a sparse GP makes the formulation amenable to stochastic variational inference (SVI),
\begin{equation}
    Y = \underbrace{F + \Phi \times B}_{\textrm{\tiny{GP with augmented kernel}}} + \underbrace{\bm{\epsilon}}_{\textrm{\tiny{noise model}}}. \hspace{2mm}
\end{equation}
%In the sections below we 
Below sections will unpack each component of the generative model and present the  formulation amenable to SVI. 

\subsubsection{Sparse GP}

% The sparse GP component is a stack of $D$ independent GPs $\{f_{d}\}_{d=1}^{D} \equiv F$ modelling each gene expression profile,

% \begin{equation}
% F = \begin{bmatrix}
% \vdots & \vdots & \ldots & \ldots & \vdots \\
% \*f_{1} & \*f_{2} & \ldots & \ldots & \*f_{d} \\
% \vdots & \vdots & \ldots & \ldots & \vdots \\
% \end{bmatrix}_{N \times D}.
% \end{equation}

% Each column of the $F$ matrix $\*f_{d} \in \mathbb{R}^{N}$ is interpreted to be an independent Gaussian process,

The sparse GP component is a stack of $D$ independent Gaussian processes $F \equiv \{\*f_{d}\}_{d=1}^{D}$, with each column modelling a particular gene expression profile,
\begin{equation}
\*f_{d} \sim \mathcal{N}(\mu_f\mathbb{I}_{N}, \sigma^{2}_{d}K_{nn}),
\end{equation}
where $\sigma^{2}_{d}$ controls the scaling of the data in column $D$. The matrix $K_{nn}$ is the usual GP covariance matrix computed with a kernel function of choice on latent inputs $X$. 

To avoid the $\mathcal{O}(N^{3})$ scaling we introduce inducing variables per gene dimension $\{\bm{u}_{d}\}_{d=1}^{D} \equiv U, \*u_{d} \in \mathbb{R}^{M}$. The $\bm{u}_{d}$'s are evaluations of $\bm{f}_{d}$ at inducing inputs $Z \in \mathbb{R}^{M \times Q}$ which lie in latent space \citep{hensman2013gaussian}. Similar to $F$, 
\begin{align}
    \*u_{d} &\sim \mathcal{N}(\*0, \sigma^{2}_{d}K_{mm}) \\
    \*f_{d} | \*u_{d} &\sim \mathcal{N}({K}_{nm}K_{mm}^{-1}\bm{u}_{d}, \sigma^2_{d}(K_{nn} - K_{nm}K_{mm}^{-1}K_{mn})). \nonumber
    \label{cond}
\end{align}

Further, $K_{nn}, K_{mm}$ and $ K_{nm}$ represent covariance matrices computed on latent inputs $(X)$, inducing inputs $(Z)$ and $X,Z$ respectively. The data noise model is given by, $\{\epsilon_{n}\}_{n=1}^{N} \equiv \bm{\epsilon} \in \mathbb{R}^{N \times D}$, where $\epsilon_{nd} \sim \mathcal{N}(0, \sigma^{2}_{y})$.

\subsubsection{Augmenting the Kernel Function}
\label{reinterpret}
This section describes a formulation where we interpret $F + \Phi B$ as a joint process (a sum of Gaussian processes). Let, $\tilde{F} = F + \Phi B$ where the columns $\tilde{\bm{f}_{d}} \sim \mathcal{N}(\mu_{f}\mathbb{I}_{N} + \Phi\zeta_{d}, K_{nn} + \nu\Phi\Phi^{T})$ are distributed identically ($\zeta_{d}$ is identical across columns $D$).
We use straight-forward identities to derive the mean and covariance for the joint process,
\begin{align}
    \mathbb{E}(\tilde{\bm{f}_{d}}) &= \mathbb{E}(\bm{f}_{d}) + \mathbb{E}(\Phi B_{d}) = \mu_{f}\mathbb{I}_{N} + \Phi\zeta_{d}\\
    \textrm{Cov}(\tilde{\bm{f}_{d}}) &= \textrm{Cov}(\bm{f}_{d}) + \textrm{Cov}(\Phi B) = K_{nn} + \nu\Phi\Phi^{T},
\end{align}
where we assume a constant mean $\mu_{f} \in \mathbb{R}$ for the $\bm{f}$ process, the design matrix $\Phi$ with covariates is specified and $\zeta_{d}$ encapsulates the mean of random effects $B$. The mean parameters $\mu_{f}$ and $\zeta_{d}$ are model hyperparameters learnt during training.  
The expression matrix $Y$ is driven by this joint process $\tilde{F}$ with columns $\bm{\tilde{f}_{d}}$ distributed as individual Gaussian processes,
\begin{align}
p(\tilde{F}) = \prod_{d=1}^{D}p(\tilde{\bm{f}}_{d}) = \prod_{d=1}^{D}\mathcal{N}(\Phi\bm{\zeta}_{d}, & K_{nn} + \nu\Phi\Phi^{T}), 
\end{align}
where the term $\nu\Phi\Phi^{T}$ contributes a linear kernel term added to the canonical kernel $K_{nn}$ driven by the kernel function described below. The parameter $\nu \in \mathbb{R}$ controls the scaling over the linear kernel.
We assume that the gene expression is affected by cell cycle and other biological effects. These effects are not known a priori, and have to be inferred from the data. For $Q$-dimensional latent
vectors $\bm{x}_{1}, \bm{x}_{2} \in \mathbb{R}^{Q}$, the kernel function designed to capture these effects is given by the product of a periodic and SE-ARD ($k_{rbf}$) kernel (each acting on non-overlapping dimensions of the $Q$ dimensional latent variable), where the periodic term models the cell cycle effect and the $k_{rbf}$ captures the other effects,
\begin{align}
    k_{\tilde{f}}(\bm{x}, \bm{x}^{\prime}) &= \sigma^{2}_{f}\exp\left\{\frac{-2\sin^{2}(|\bm{x}_{1} - \bm{x}_{1}^{\prime}|/2)} {l_{1}^{2}} \right\} 
    \times\exp\left\{-\sum_{q=2}^{Q}\frac{(\bm{x}_{q} - \bm{x}^{\prime}_{q})^{2}}{2l_{q}^{2}}\right\} + \nu\Phi\Phi^{T} \\
    &= k_{per} \times k_{rbf} + k_{lin}.
    \label{kernel}
\end{align}
The hyperparameters for the mean and covariance are collected in $\bm{\theta} = \{\sigma^{2}_{f},\{l_{q}\}_{q=1}^{Q} \nu, \mu_{f}, \zeta_{d}\}.$

Let $\tilde{K}_{nn} = K_{nn} + \nu\Phi\Phi^{T}$; it is important to clarify how the cross-covariance terms $\tilde{K}_{nm}$ and $\tilde{K}_{mm}$ are computed in the augmented formulation. The inducing inputs/locations $Z$ now have to incorporate additional dimensions to account for linear kernel term. Hence, they have dimension $M \times (Q + \textrm{col. dim}(\Phi))$. One can think of this as a partitioning of $Z = [\underbrace{Z_{per}|Z_{rbf}}_{Q-dims}|Z_{lin}]$ where the first $Q$ columns feed into $k_{per} \times k_{rbf}$ and the columns corresponding to the col. dim($\Phi$) are used to compute the linear kernel term,
\begin{align}
    \tilde{K}_{nm} &= K_{nm} + \nu\Phi Z_{lin}^{T} \\
     \tilde{K}_{mm} &= K_{mm} + \nu Z_{lin}Z_{lin}^{T} 
\end{align}
where $K_{nm}$ and $K_{mm}$ have been computed on the first $Q$ columns $Z_{per}|Z_{rbf}$.

\subsubsection{Variational Formulation}
The joint probabilistic model is given by,
\begin{align}
    p(Y, \tilde{F}, \tilde{U}) &= p(Y|\tilde{F}, \Sigma)p(\tilde{F}|\tilde{U})p(\tilde{U}|Z) \\
    &\hspace{-15mm}= \mathcal{N}(Y|F  + \Phi\bm{\zeta},\Sigma)\mathcal{N}(\tilde{F}|\tilde{K}_{nm}\tilde{K}_{mm}^{-1}\tilde{U}, \tilde{Q}_{nn}) 
    \mathcal{N}(\tilde{U}|0, \tilde{K}_{mm}), \nonumber
   % &= \prod_{n=1}^N \prod_{d=1}^D \mathcal{N}(y_{n,d}| \bm{f}_{d}(\bm{x}_{n}) + (\Phi \bm{\zeta})_{n,d}, \omega^2_n\sigma^{2}_{d}) \displaystyle \prod_{d=1}^{D}\mathcal{N}(\bm{f}_{d}| \tilde{K}_{NM}K_{MM}^{-1}\bm{u}_{d}, {\sigma^{2}_{d}}Q_{NN})\displaystyle \prod_{d=1}^{D}\mathcal{N}(\bm{u}_{d}| 0, {\sigma^{2}_{d}}K_{MM}) \\   &\times \displaystyle \prod_{p=1}^{P}\prod_{d=1}^{D} \displaystyle \prod _{n=1}^N \mathcal{VM}(x_{n,1}|c_n,\kappa)\prod _{q=2}^Q \mathcal{N} (x_{n,q}| 0, 1) \nonumber
\end{align}
where $\tilde{Q}_{nn} = \tilde{K}_{nn} - \tilde{K}_{nm}\tilde{K}_{mm}^{-1}\tilde{K}_{mn}$. The marginal likelihood $p(Y)$ is intractable and we resort to variational inference over $\tilde{U}$ and point estimation for latents $X$, model hyperparameters $\bm{\theta}$ and variational parameters $Z$.
The posterior over unknowns is approximated by the variational distribution,
\begin{align}
p(\tilde{F}, \tilde{U} | Y) &\approx p(\tilde{F}| \tilde{U}, X )q(\tilde{U}|Z)\\
&= \prod_{d=1}^{D}[p(\tilde{\bm{f}_{d}}|\tilde{\bm{u}}_{d}, X)q(\tilde{\bm{u}}_{d}|Z)],
\end{align}
where $p(\tilde{\bm{f}_{d}}|\tilde{\bm{u}}_{d}, X)$ is as in eq. \ref{cond} and $q(\tilde{\bm{u}}_{d}|Z) = \mathcal{N}(\bm{m}_{d}, S_{d})$ where $S_{d}$ is a dense $M \times M$ covariance matrix. $\{\bm{m}_{d}, S_{d} \}_{d=1}^{D}$ are global variational parameters.

We use the standard stochastic variational evidence lower bound (ELBO) which preserves factorisation across the data-points $N$, and the final bound used to generate the results takes the form, 
\begin{align}
     \log p(Y) &\geq \sum_{n=1}^{N}\sum_{d=1}^{D} \Big\{\log \mathcal{N}(y_{n,d}|\tilde{k}_{n}\tilde{K}^{-1}_{mm}\bm{m}_{d} ,\sigma^{2}_{y}) - \dfrac{q_{n,n}}{2\sigma^{2}_{y}} 
     -\dfrac{\textrm{Tr}(S_d\Lambda_{n})}{2\sigma^{2}_{y}} \Big\} - \textrm{KL}(q(\tilde{U}|Z)||p(\tilde{U}|Z)),
\end{align}
where $\Lambda_{n} = \tilde{K}_{mm}^{-1}\tilde{k}_{mn}\tilde{k}_{nm}\tilde{K}^{-1}_{mm}$,  $\tilde{k}_{n}$ is the $n^{th}$ column of $\tilde{K}_{mn}$, and $q_{n,n}$ is the $n^{th}$ diagonal entry of the matrix $Q_{nn}$ defined earlier. Note that this is the %exactly the
 multi-output lower bound used by \citet{hensman2013gaussian} in a regression context and by \citet{lalchand2022generalised} in an unsupervised context with variational inference over $X$, whereas in  here we learn a point-estimate for $X$. We include our algorithm in Appendix \ref{algo}.   

\subsection{Data}

The innate immunity dataset (\citet{kumasaka2021mapping}) comprises gene expression profiles of 22,188 primary dermal fibroblasts from 68 donors exposed to two stimulants to mimic innate immune response: (1) the synthetic dsRNA Poly(I:C) for primary antiviral and inflammatory responses, and (2) Interferon-beta (IFN-beta) for secondary antiviral response. Cells were processed in 128 plates representing the technical batch. The COVID dataset (\citet{stephensonSinglecellMultiomicsAnalysis2021}) consists of gene expression profiles of peripheral blood mononuclear cells from a cohort of 107 patients with varying severities of COVID-19 and 23 healthy individuals. Samples have been collected and processed at three different sites (Sanger, Cambridge, Newcastle).
We sampled 5\% (32,368 cells) and 10\% (54,941 cells) of the full dataset for dimensionality reduction comparison analysis (Fig. ~\ref{fig:covid_celltypes_batch}-~\ref{fig:lv_interpret}) and latent severity analysis, respectively. 
%For the dimensionality reduction comparison analysis (Fig. ~\ref{fig:covid_celltypes_batch}-~\ref{fig:lv_interpret}), we sampled 5\% of cells in the dataset (32,368 cells). For the latent severity analysis we sampled 10\% of cells in the dataset (54,941 cells). 
For the encoder dimensionality reduction experiment, we used the full dataset (647,366 cells).

For all datasets, their gene expression counts were normalized to sum up to 10,000 counts per cell and $\log(x+1)$ transformed as per the  standard practice of preprocessing. For dimensionality reduction analysis, we selected the 5000 most highly variable genes, as in Scanpy \citep{wolfSCANPYLargescaleSinglecell2018}.

\subsection{Model Inference}

Across all datasets, our model inference  was done by initializing latent variables by the $Q$-largest principal components of the data, and utilizing expression of S, G2 and M state gene markers to initialise the cell cycle latent variable. (See Appendix \ref{init} for a detailed discussion on initialization). The models were trained using learning rates of $5*10^{-3} \text{ to } 10^{-2}$ for $\sim$50 epochs using batch sizes of $\sim$200-700 data points for SVI. We observe that, training the model excluding the latent variables (i.e. the sparse GP) for few epochs with high learning rate before jointly optimising all parameters with a lower learning rate leads to good model performance.

\section{Results}

\subsection{Reproducing Innate Immunity Analysis}

\begin{figure}
\centering
   \includegraphics[width=\textwidth]{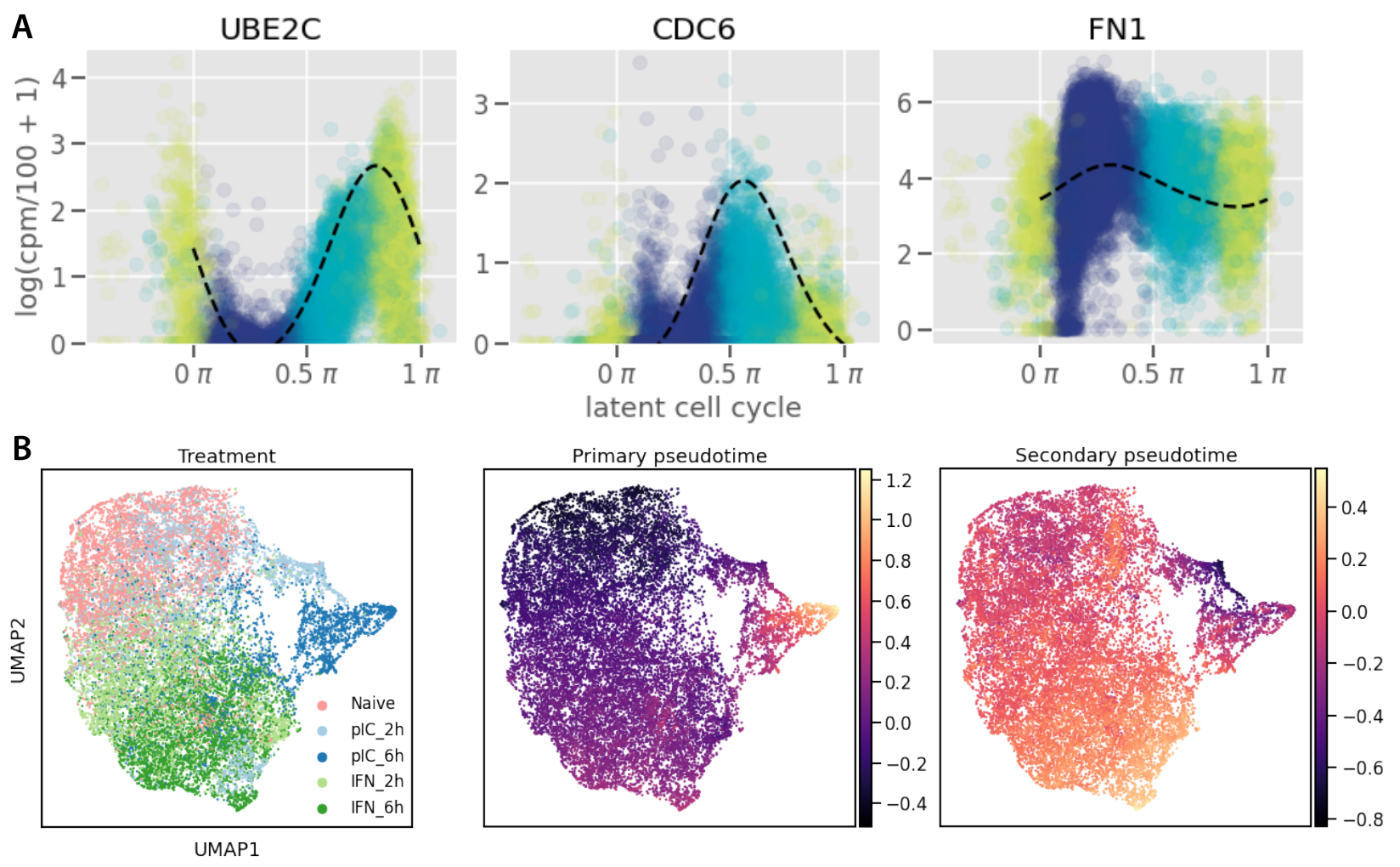}
    \caption{
        (A) Scatterplots of scaled expression of known cell cycle genes (UBE2C, CDC6, FN1). The dotted curves show the posterior mean expression estimates %of expression %levels 
        by GPLVM. Points are colored by predicted cell cycle phase.
        (B) UMAP embedding of innate immunity dataset latent space learnt by GPLVM. Cells are colored by treatment condition (pIC: Poly(I:C), IFN: interferon-beta) (left), primary immune response pseudotime latent variable (center) and secondary immune response pseudotime latent variable (right), as described by \citet{kumasaka2021mapping}. Latent dimensions for %capturing hidden
        hidden technical effects (batch and plate border effects) were excluded for UMAP embedding. 
    }
    \label{fig:ipscs_umaps}
\end{figure}

We first tested our framework on a dataset previously analysed with the additive GPLVM framework (eq. \ref{nats}) used in \citet{kumasaka2021mapping}, to demonstrate the ability to capture validated latent variables 
at a fraction of the speed. Our run was about nine times faster (30 min  compared to a baseline of 4.5 hours) and is able to reproduce key results described in \citet{kumasaka2021mapping}. Our GPLVM model can disentangle batch effects and the gene expression signature related to the cell cycle (Fig. ~\ref{fig:ipscs_umaps_all}), which is captured with the use of the periodic kernel (\cref{fig:ipscs_umaps}A). In addition, latent variables recover the two main axes of biological variation in this data: (1) response to Poly(I:C) treatment, and (2) response to IFN-beta treatment from the naive state, mimicking primary and secondary immune response (as known as response pseudotime) (Fig. ~\ref{fig:ipscs_umaps}B). 

% - Runtime analysis (R code VS this code on full data or subsets by cells)
% - Scatterplot of LV for primary immune response vs LV for secondary immune response separating cells by condition (Naive/pIC/IFN)

% \begin{figure}
% \centering
%   \includegraphics[width=0.5\textwidth]{example-image-a}
%     \caption{
%         Run time (\textit{y}-axis) as a function of number of cells for the sparse GPLVM model implemented in \citet{kumasaka2021mapping} and model implemented in this work.
%     }
%     \label{fig:ipscs_time}
% \end{figure}

\subsection{Modelling COVID Severity with GPLVM}
\label{severity}

The ability to correct for confounders with random effects makes the GPLVM an amenable model for analysis of cohort-level scRNA-seq datasets. 
% We applied our model on the COVID dataset encoding the site of collection and sample ID in the design matrix for random effects. 
% We show performance of the model on tasks such as the  identification of cell type clusters and accounting for batch effects, and compare to alternative dimensionality reduction approaches in Appendix \ref{dimred}. 

% \section{Dimensionality Reduction on Disease Cohort Dataset}
% \label{dimred}

\begin{figure*}
\centering
   \includegraphics[width=1.1\textwidth]{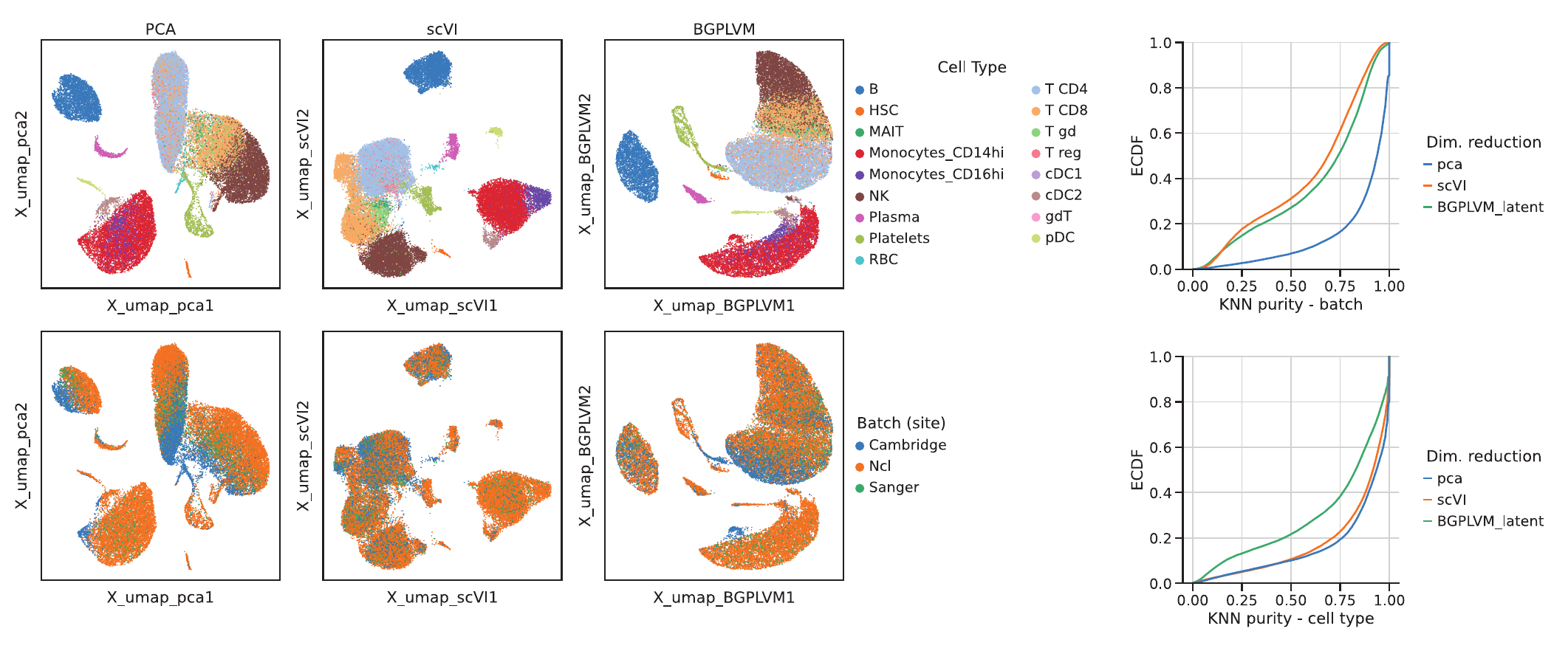}
    \caption{
        Cell type clustering and batch correction task on COVID dataset
        (a) UMAP embedding of cells generated from latent dimensions (excluding batch and cell cycle variables) learnt by PCA (left), scVI (middle), and scalable GPLVM with random effects (right). Cells are colored by annotated cell type (top) and batch (bottom)
        (b) Cumulative Distribution of k-nearest neighbor (KNN) purity of cell type labels (top) and batch labels (bottom) in each latent space.
    }
    \label{fig:covid_celltypes_batch}
\end{figure*}

\begin{figure}
\centering
    \includegraphics[scale=0.37]{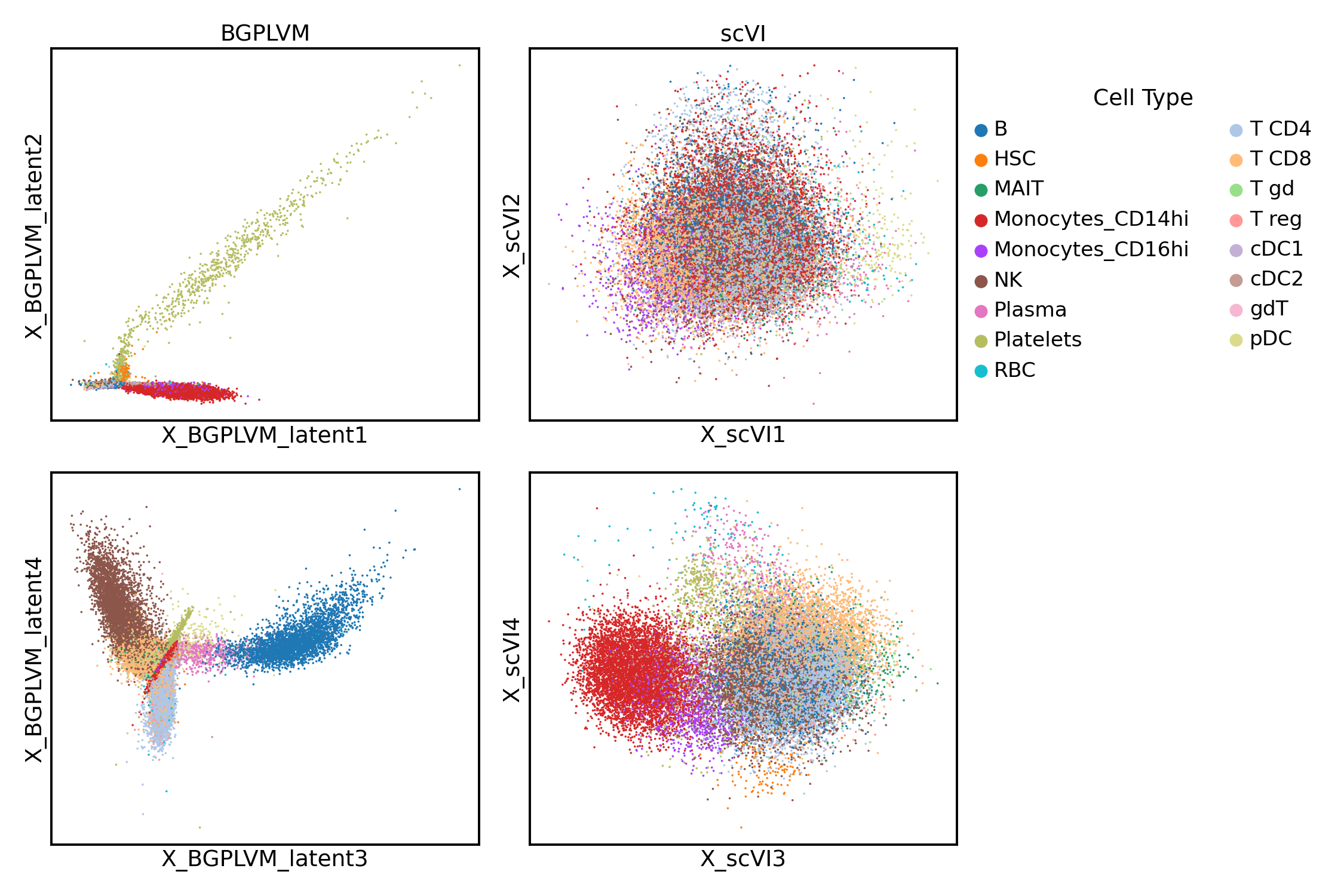}
    \caption{
        Scatterplots of four latent dimensions learnt with GPLVM (left) and scVI (right). Points are colored by cell type.
    }
    \label{fig:lv_interpret}
\end{figure}

We first assessed the ability to identify cell type clusters while accounting for batch effects. We trained our model by encoding the site of collection and sample ID in the design matrix for random effects. We compared our model to PCA (dimensionality reduction without batch correction) and a conditional VAE (cVAE) for dimensionality reduction (scVI \citet{lopezDeepGenerativeModeling2018}) trained by encoding the same technical conditions as categorical covariates and the cell cycle initialization as a continuous covariate.
We visualize the results by UMAP embedding of latent spaces and we quantify mixing between cell labels by measuring the fraction of k-nearest neighbors of a cell (k=100) harboring the same label (KNN purity). 
While the scVI model performs better on the cell type clustering task, our model was able to recover broad cell type clusters and minimize differences driven by batch effects (Fig. ~\ref{fig:covid_celltypes_batch}). In addition, the GPLVM latent dimensions capture interpretable biological differences between cells, which are not discernible using the latent dimensions from cVAE based models. For example, we find latent variables that capture the cell type-specific gene expression signatures (e.g. Platelets in LV1 and LV2, separation between B cell and NK/T cell lineage in LV3) (Fig. ~\ref{fig:lv_interpret}, top). Moreover, \citet{stephensonSinglecellMultiomicsAnalysis2021} found increased differentiation of megakaryocyte (MK) cells (megakaryopoiesis) and platelet activation associated to COVID. One of the GPLVM latent variables was strongly associated with the expression of megakaryopoiesis genes in platelets (Fig. ~\ref{fig:lv_interpret}, middle) and this variable separates platelets from healthy and asymptomatic patients from those of patients with severe disease (Fig. ~\ref{fig:lv_interpret}, bottom). In contrast, none of the scVI latent variables were significantly associated with this signature.  

\begin{figure}
\centering
   \includegraphics[width=0.7\textwidth]{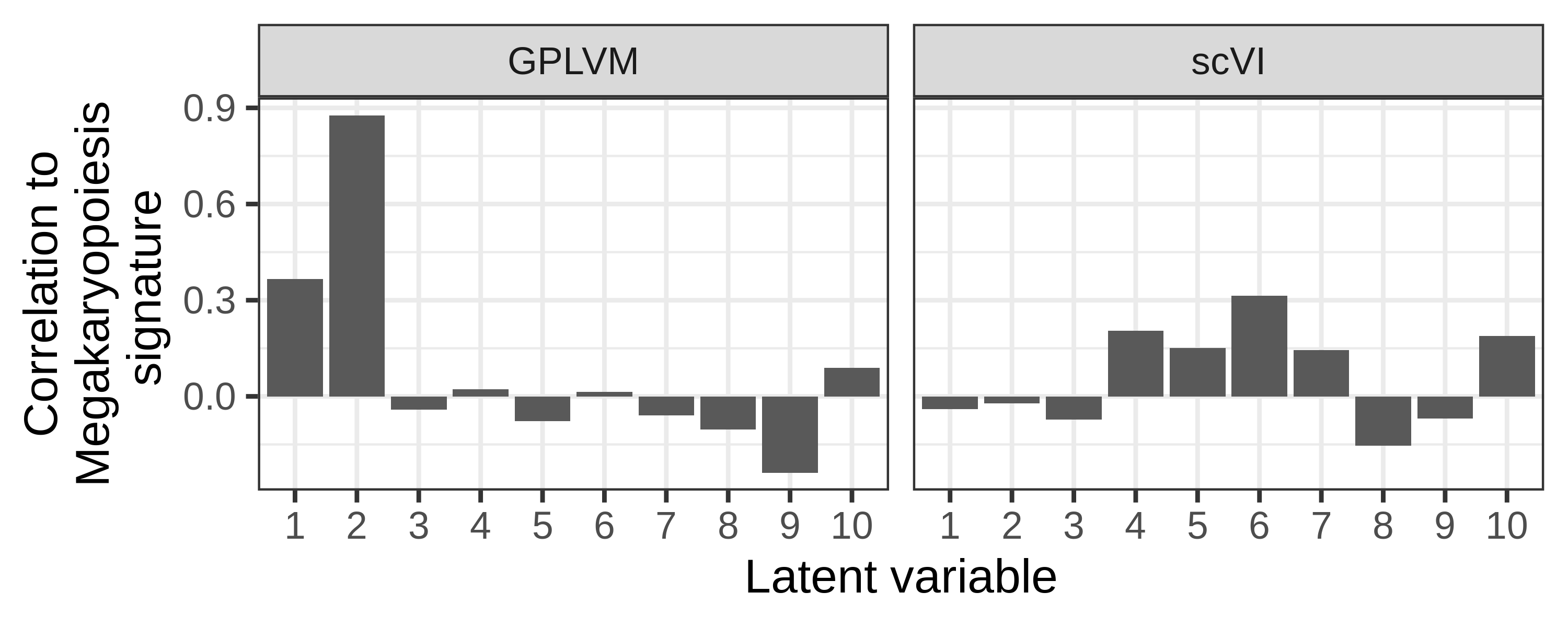}
   \includegraphics[width=0.7\textwidth]{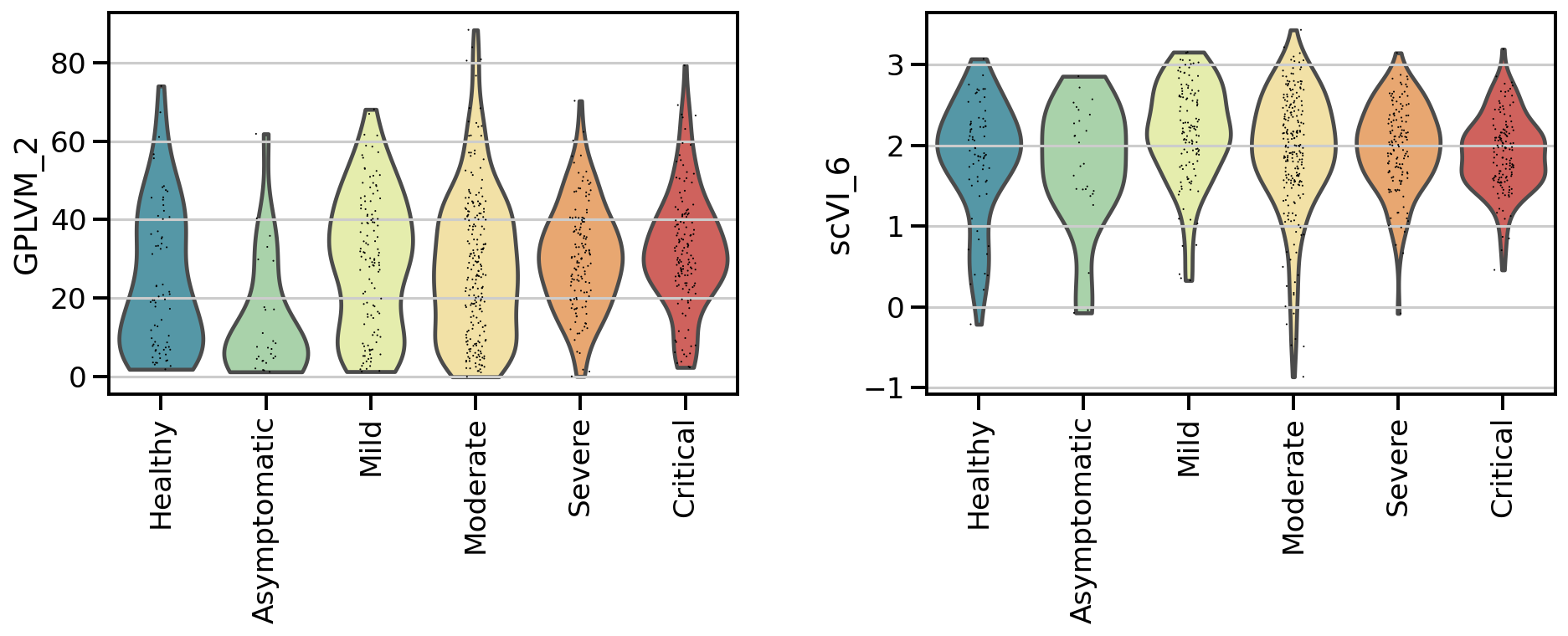}
    \caption{
       (middle) Barplot of Pearson correlation between latent variable values and megakaryopoiesis gene expression signature. The expression of 7 marker genes (TUBB1, PPBP, PF4, TREML1, GP9, MYL9, F13A1) was aggregated by taking the average expression of these genes subtracted with the average expression of a background set of genes with matched expression across cells. 
       (bottom) Distribution of values for LVs (\textit{y}-axis) in platelet cells grouped by disease severity (\textit{x}-axis). LVs most associated with megakaryopoiesis signature with GPLVM (left) and scVI (right) are shown.
    }
    \label{fig:mk_sig}
\end{figure}

\begin{figure}
\centering
   \includegraphics[scale=0.63]{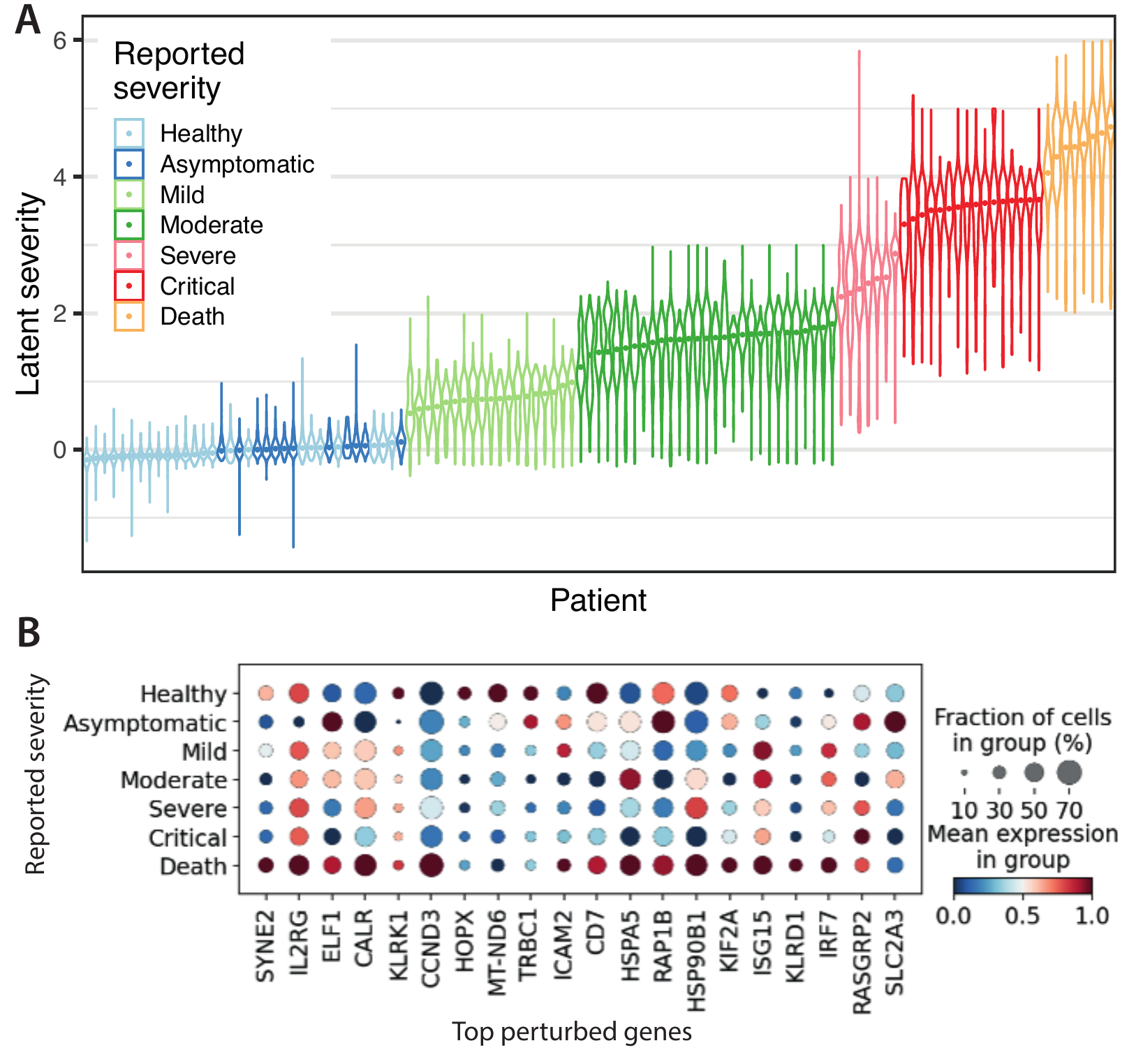}
    \caption{
        (A) Violin plots of latent COVID severity inferred by GPLVM for each patient in COVID study. Patients are ordered by the median latent severity (point) and colored by reported disease severity (worst clinical outcome).
        (B) Average expression of genes identified as variable across severities in perturbation experiment. 
    }
    \label{fig:severity}
\end{figure}

One powerful feature of the GPLVM is its ability to incorporate covariates of interest in the latent dimensions to generate a meaningful dimensionality reduction. In cohort studies, patients are often grouped by disease severity using qualitative assessments, which can be inaccurate and do not fully reflect the true spectrum of observed severities. Incorporating a latent variable in the GPLVM to capture disease severity %could allow 
supports better patient stratification and detection of gene expression signatures predictive of clinical outcome. We demonstrate this feature by training the GPLVM on 54,941 cells from the COVID dataset with an additional continuous latent variable initialized on disease severity (converted to an integer encoding). After training, the GPLVM refines disease severity assignment in a meaningful and cell-specific manner:  cells from healthy individuals and asymptomatic infected patients are placed closer together in latent severity than cells from symptomatic infections (Fig. ~\ref{fig:severity}A). Symptomatic cases are placed along a more fine grained spectrum of disease severity, which could be helpful to better classify cases. For example, in the moderate cases, we find that the predicted severity is significantly correlated with the days since onset of symptoms ($R^2 = 0.25, \text{p-val} < 2.2e-16$, Fig. ~\ref{fig:moderate_severity}). In addition, we can use the generative model to predict genes that are most affected by changes in disease severity, while minimizing the effect of confounders.
This is done by inputting a uniformly varying disease severity (with all other inputs set to be constant) to the sparse GP  %Gaussian process 
generative model, which outputs predicted gene expression for different levels of severity.
Amongst the top 20 genes most affected by changes in severity, we find genes associated with interferon signalling (IRF7, ISG15), which is a recognized signature of COVID disease (\citet{stephensonSinglecellMultiomicsAnalysis2021}), markers of viral entry (HSPA5), as well as several cell type markers (e.g. KLRK1, TRBC1, CD7, KLRD1), potentially reflecting shifts in cell abundances in response to infection (Fig. ~\ref{fig:severity}B).

% \begin{figure*}
% \centering
%   \includegraphics[width=0.5\textwidth]{example-image-a}
%     \caption{
%         Correction of cell cycle effect in latent embedding
%         (a) UMAP embedding of cells generated from latent dimensions learnt by scVI and scalable GPLVM. Colored by annotated MKI67 expression (log(x+1) normalized counts).
%         (b) Cumulative Distribution of k-nearest neighbor (KNN) purity of proliferating cells in each latent space.
%     }
%     \label{fig:covid_prolif}
% \end{figure*}

\section{Scaling up to Large Scale Dataset with Amortized Inference}
\label{encoder}
With datasets now collecting up to millions of cells, our SVI formulation can be combined with strategies such as amortized inference for application to massive scale datasets. We demonstrate scalability to very large datasets on the full COVID infection dataset ($>$ 600k cells)
using neural network based encoders as variational approximations for GPLVM latent variables.

In this version of the  GPLVM, we variationally integrate out the latent variables $X$, with a standard normal prior $p(X)$, the mean and variance of the variational Gaussian distribution are parameterised as outputs of individual neural networks $G_{\phi_{1}}$ and $H_{\phi_{2}}$ with network weights $\phi_{1}$ and $\phi_{2}$. The network weights are shared across all the cells ($N$) enabling amortised learning \citep{bui2015stochastic}. %The key property of this parameterisation is that it learns a dense covariance matrix (parameterised through a factorization) per data-point thereby capturing correlations across dimensions (per latent point) in latent space. 
\begin{equation}
q(X) = \prod_{n=1}^{N}\mathcal{N}( \bm{x}_{n}; G_{\phi_{1}}(\bm{y}_{n}), \text{diag}(H_{\phi_{2}}(\bm{y_{n}}))) 
\label{nn_bc}
\end{equation}
This function is usually referred to as the back-constraint or an encoder and its parameters are \textit{global}, i.e. shared between all the data points and updated in each iteration. This allows for fast amortised inference as it precludes from learning free-form mean and standard deviations for each individual data point (cell). In other words, the number of variational (or model) parameters does not grow with the number of data points. We choose to model the covariance of each $\bm{x}_n$ a diagonal covariance as this leads to better behaviour while training (as opposed to when a dense covariance is allowed).

\begin{figure*}
\centering
   \includegraphics[scale=0.4]{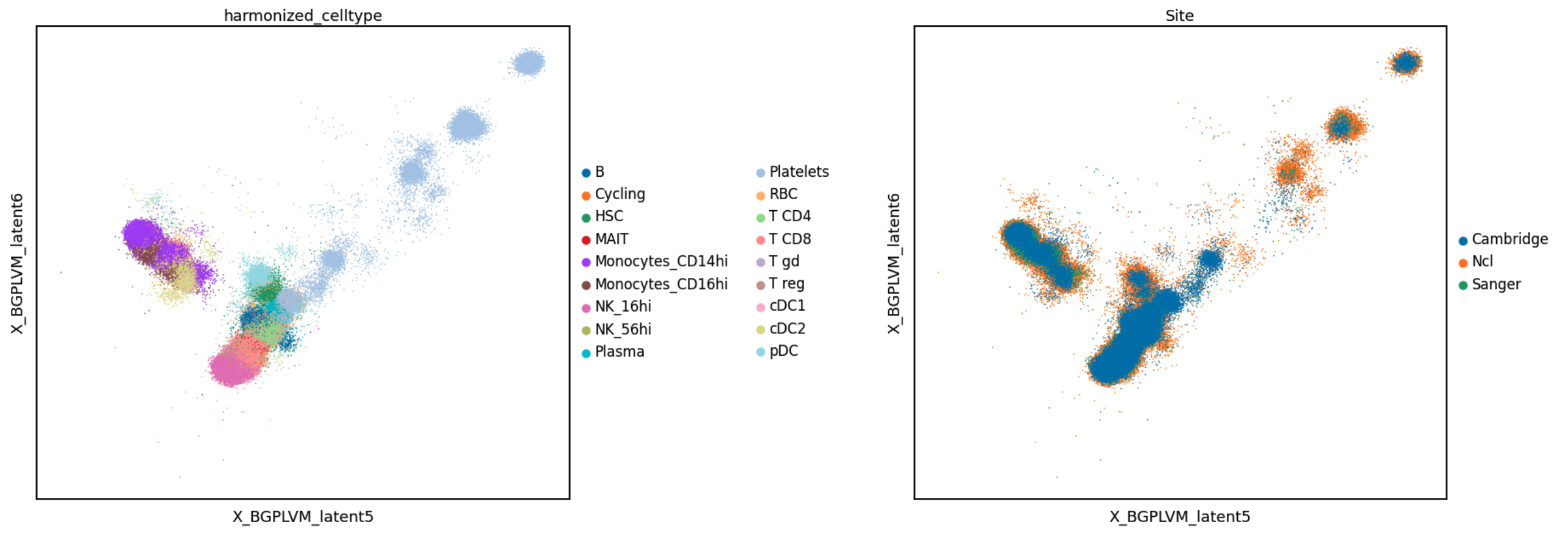}
    \caption{
        Top two latent dimensions (selected by inverse lengthscales) learned using a GPLVM fit on the full COVID dataset colored by cell types and site information.
    }
    \label{fig:covid_full_lvs}
\end{figure*}

\cref{fig:covid_full_lvs} shows results of a run using this variational set-up on the full COVID dataset comprising of over 600K cells, demonstrating that our model can be scaled up to very large datasets while learning meaningful biological representations such as broad cell types.

\section{Summary}
This work introduces an augmented kernel function to jointly model known and unknown technical and biological covariates in scRNA-seq datasets, with a formulation amenable for scale-up using stochastic variational inference. 
% This allows a \ed{n}x speed up compared to previous GPLVM models accounting for confounders, while recovering the same latent dimensions of biological and technical variation. 
With datasets now collecting up to millions of cells, our SVI formulation can be combined with strategies such as amortized inference for application to massive scale datasets. We demonstrate scalability to very large datasets on the full COVID infection dataset ($>$ 600k cells) % in Appendix ~\ref{encoder}, 
using neural network based encoders as variational approximations for GPLVM latent variables. We envision this model will have broad application to population-scale scRNA-seq studies, where complex confounders and inter-individual variability need to be accounted for, as shown in our analysis on COVID severity. 

\section{Code availability}
The model and code to reproduce the analyses presented here is available at \href{`https://github.com/InfProbSciX/gplvm_scrna`}{\url{https://github.com/InfProbSciX/gplvm_scrna}}.Our software is implemented in GPyTorch and can be integrated into existing libraries of probabilistic single-cell data analysis  (\href{https://scvi-tools.org/}{https://scvi-tools.org/}) for accessibility to the wider community.

\section{Acknowledgements}
This work was supported by the Wellcome Trust grant reference: 221052/Z/20/Z, 221052/A/20/Z, 221052/B/20/Z, 221052/C/20/Z, and 221052/E/20/Z.

\bibliography{ref}
\bibliographystyle{plainnat}

%%%%%%%%%%%%%%%%%%%%%%%%%%%%%%%%%%%%%%%%%%%%%%%%%%%%%%%%%%%%%%%%%%%%%%%%%%%%%%%
%%%%%%%%%%%%%%%%%%%%%%%%%%%%%%%%%%%%%%%%%%%%%%%%%%%%%%%%%%%%%%%%%%%%%%%%%%%%%%%
% APPENDIX
%%%%%%%%%%%%%%%%%%%%%%%%%%%%%%%%%%%%%%%%%%%%%%%%%%%%%%%%%%%%%%%%%%%%%%%%%%%%%%%
%%%%%%%%%%%%%%%%%%%%%%%%%%%%%%%%%%%%%%%%%%%%%%%%%%%%%%%%%%%%%%%%%%%%%%%%%%%%%%%
\newpage
\appendix
% \begin{appendices}
\onecolumn

\beginsupplement
\section*{Supplementary Material}

\section{Graphical Model}

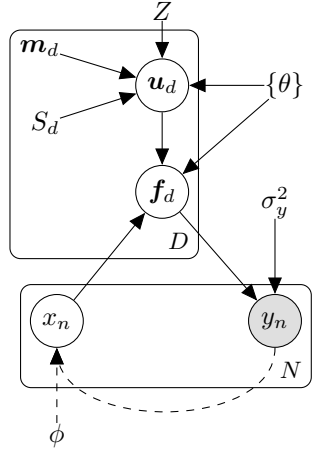
\begin{figure}[h]
\centering
   \input{graphical_additive_kernel}
    \caption{The probabilistic graphical model presented in this work. The unshaded circles denote latent variables to be inferred. The point estimate variational and model parameters are denoted by just their symbol. The dataset $Y$ is informed by the sparse GPs, random effects and noise term. The latent variables $x_{n}$ are inferred through a neural network encoder with weights $\phi$, alternatively, just learnt as a point estimate.}
    \label{fig:graph}
\end{figure}

\cref{fig:graph} depicts a graphical model of the sparse Gaussian process used in this paper.

\section{Random Effects}

The matrix $B$ models random effects across $P$ categorical and $J$ continuous variables. Let $c_p$ denote the number of levels of the $p^{th}$ categorical variable. The design matrix $\Phi$ has $N$ rows and column dimension $\sum_{p=1}^{P}c_{p} + J$ (there is a column per level per categorical variable and one for each continuous variable). Each cell (row of $Y$, $\bm{y}_{n}$) is associated with a particular level within each of the $P$ categories \footnote{For example, if the two categories are gender (with 2 levels) and ethnicity (with 8 levels) then each cell $\bm{y}_{n}$ is associated with one of two levels in the gender category and one of eight levels in ethnicity category.}. The $n^{th}$ row of the design matrix $\Phi$ would have a 1 in the columns to reflect membership to a particular category and reals in the columns which reflect co-efficients for the continuous covariates. The row dimension of $B$ is $\sum_{p}c_{p} + J$. The dimension of $\Phi B$ is thus, $N \times D$. %One can think of $B$ as a vertical stack of random effect terms $B \equiv \{B_{p}\}_{p=1}^{P}$ and $B_{p} \in \mathbb{R}^{c_{p} \times D}$ and $\Phi$ as a horizontal stack of binary matrices $\Phi \equiv \{\Phi_{p}\}_{p=1}^{P}$  and $\Phi_{p} \in [0,1]^{N \times c_{p}}$. 
\subsection{Additive Framework: Non-factorisability}
\label{nonfactor}
In sparse GP additive model the final likelihood of the data $Y$ can be expressed as, 
\begin{align}
      p(Y| F, B,\sigma^{2}_{y}) &=  \prod_{n=1}^N \prod_{d=1}^D \mathcal{N}(y_{n,d}| \bm{f}_{d}(\bm{x}_{n}) + (\Phi B)_{n,d}, \sigma^{2}_{y})\\
      &=\prod_{d=1}^D\mathcal{N}(\bm{y}_{d}| \bm{f}_{d}(X) + (\Phi B)_{:,d}, \sigma^{2}_{y}\mathbb{I}_{N})
\end{align}\\
A common approach is to integrate out the random effects term $B$ which we can do in closed form due to the linear dependence. We assume independence between the columns of $B$, with each $B_{d} \sim \mathcal{N}(\zeta_{d}, \Delta_{d})$. Let $\Sigma$ denote a diagonal matrix with entries $\sigma^{2}_{y}$ on the diagonal.\\
\begin{align}
      p(Y| F, \sigma^{2}_{y})&= \int \prod_{d=1}^{D} p(\bm{y}_{d}|\bm{f}_{d},B_{d})p(B_{d})dB_{d} \\
    &= \int \prod_{d=1}^D\mathcal{N}(\bm{y}_{d}| \bm{f}_{d}(X) + (\Phi B)_{:,d}, \sigma^{2}_{y}\mathbb{I}_{N})p(B_{d})dB_{d} \nonumber\\
     &= \int \prod_{d=1}^D \mathcal{N}(\bm{y}_{d}|\bm{f}_{d}(X) + (\Phi B)_{:,d}, \Sigma)\mathcal{N}(\zeta_{d}, \Delta_{d})dB_{d} \nonumber \\
     &= \prod_{d=1}^D\mathcal{N}(\bm{y}_{d}|\bm{f}_{d}(X) + \Phi\zeta_{d}, \Sigma + \Phi\Delta_{d}\Phi^{T}) \nonumber
\end{align}
where $\zeta_{d}$ is a vector of size $\sum_{p=1}^{P}c_{p} + J$ and $\Delta_{d}$ is a matrix of size $(\sum_{p=1}^{P}c_{p} + J) \times (\sum_{p=1}^{P}c_{p} + J)$. Integrating out the random effects $B$ from the likelihood introduces a dense covariance matrix $\Phi\Delta_{d}\Phi^{T}$ in the likelihood.
%where we succintly express the distribution over random effects $p(B)$ by accumulating the means and covariance matrices in $\zeta$ and $\Delta$ is a matrix of size ${\sum c_{p}D \times \sum c_{p}D}$. 
The non-factorisable likelihood (in each of the $D$ output dimensions) prevents the application of stochastic mini-batching as we need the whole dataset in memory to compute a single gradient step. Instead, we present an alternative formulation in section \ref{reinterpret} which subsumes the random effects within the kernel function and preserves factorisability.

%\section{Deriving the Evidence lower bound (ELBO)}
%\label{deriv}

\section{Inference Algorithm}
\label{algo}
\begin{algorithm*}
\SetAlgoLined
\label{algorithm}
\footnotesize
\caption{Stochastic Variational Inference for Scalable GPLVM with random effects }
\vspace{2mm}
\textbf{Input:} ELBO objective $\mathcal{L}$, gradient based optimiser \texttt{optim()}, training data $Y = \{\bm{y}_{n}\}_{n=1}^{N}$ \\ 
Initial model params: \\
\quad $\bm{\theta}$ (covariance hyperparameters for GP mappings $\bm{f}_{d}$), \\
\quad $\sigma^{2}_{y}$ (variance of the noise model), \\
\quad $X \equiv \{\bm{x}_{n}\}_{n=1}^{N}$ (point estimates for latent covariates)\\
%\quad $\zeta$ (matrix for mean parameters of covariates) \\
%\quad $\Delta$ (diagonal matrix for variance parameters of covariates) \\
 \vspace{1mm}
Initial variational params: \\
\quad $Z \in \mathbb{R}^{M \times (Q + col.dim(\Phi))}$ (inducing locations), \\
%\quad $\phi = \{\mu_{n}, s_{n} \}_{n=1}^{N}$  (local variational params per latent point $\bm{x}_{n}$, alternatively weights of the neural network encoder),\\
\quad $\lambda = \{m_{d}, S_{d} \}_{d=1}^{D}$ (global variational params for inducing variables per dimension $\bm{u}_{d}$), \\
%\quad $\eta = \{(\bm{r}_{p})_d, (h_{p}) \}_{p=1}^{P}{}_{d=1}^D$ (global variational params for random effects $B$) \\
\vspace{2mm}
\While{not converged}{ 
\begin{itemize}
  \setlength{\itemsep}{0pt}
   \item Choose a random mini-batch ${Y}_{B} \subset Y$. \\
   %\item Sample $J$ samples from the noise distribution $\epsilon^{(j)} \sim \mathcal{N}(0,\mathbb{I}_{Q})$. \\
   \item Form a mini-batch estimate of the ELBO: \\
   \qquad $\mathcal{L}(Y_{B}) = \dfrac{N}{B}\left(\sum_{b}\sum_{d}\mathcal{L}_{b,d} \right) - \sum_{d}\textrm{KL}(q(\bm{u}_{d})||p(\bm{u}_{d}|Z))$\\
   \item Gradient step: $Z, \bm{\theta}, \sigma^{2}_{y}, \lambda, X_{B} \equiv \{\bm{x}_{n}\}_{n=1}^{B} \longleftarrow $ \texttt{optim}$(\mathcal{L}(Y_{B}))$
\end{itemize}
}
\Return{$ \bm{\theta}, \sigma^{2}_{y}, Z, \lambda, X$}
\end{algorithm*}

\section{Initialization and Training Configurations}
\label{init}

In the experiments presented here, we use informative initializations for the GPLVM latent variables. 

We use the expression of S, G2 and M state (cell cycle) gene markers to initialize the cell cycle effect latent variable modelled with periodic covariance (Note: S,G2, M denote the synthesis phase, cell growth phase and cell division phase in the cell cycle, respectively). While alternative analysis workflows suggest to regress out this signal before dimensionality reduction, the GPLVM provides a way to inject prior knowledge regarding this (through the periodic kernel and latent dimension initialisation). Overall, we can consider the GPLVM as a unified probabilistic framework that gives flexibility and consistency to analyse multiple covariates in the data. We initialize the other latent variables by top principal components of the data, which turn out to be interpretable in some cases, for example, by corresponding to known batch and technical effects in the data.

We find that informative initializations help model performance.  Our software allows flexibility for incorporating various combinations of known and latent variables to be used with the periodic, linear and smooth components of our covariance function. For example, spatial coordinates corresponding to well location of sequencing plates can be input to the smooth component of the model as a known covariate if it is expected to reduce confounding effect. Additionally, when neural-network based encoders are used (as described in \cref{encoder}), a cell cycle initialization vector can be passed as a known covariate to the periodic component of our kernel, somewhat mirroring to the process used in conditional VAEs.

We use a number of inducing variables strictly greater than the number of covariates (Col. dim ($\Phi$)) used for the linear kernel as this is the minimum number of data points needed to define a hyperplane in Col. dim($\Phi$) dimensions.

\begin{figure}
\centering
   \includegraphics[width=0.8\textwidth]{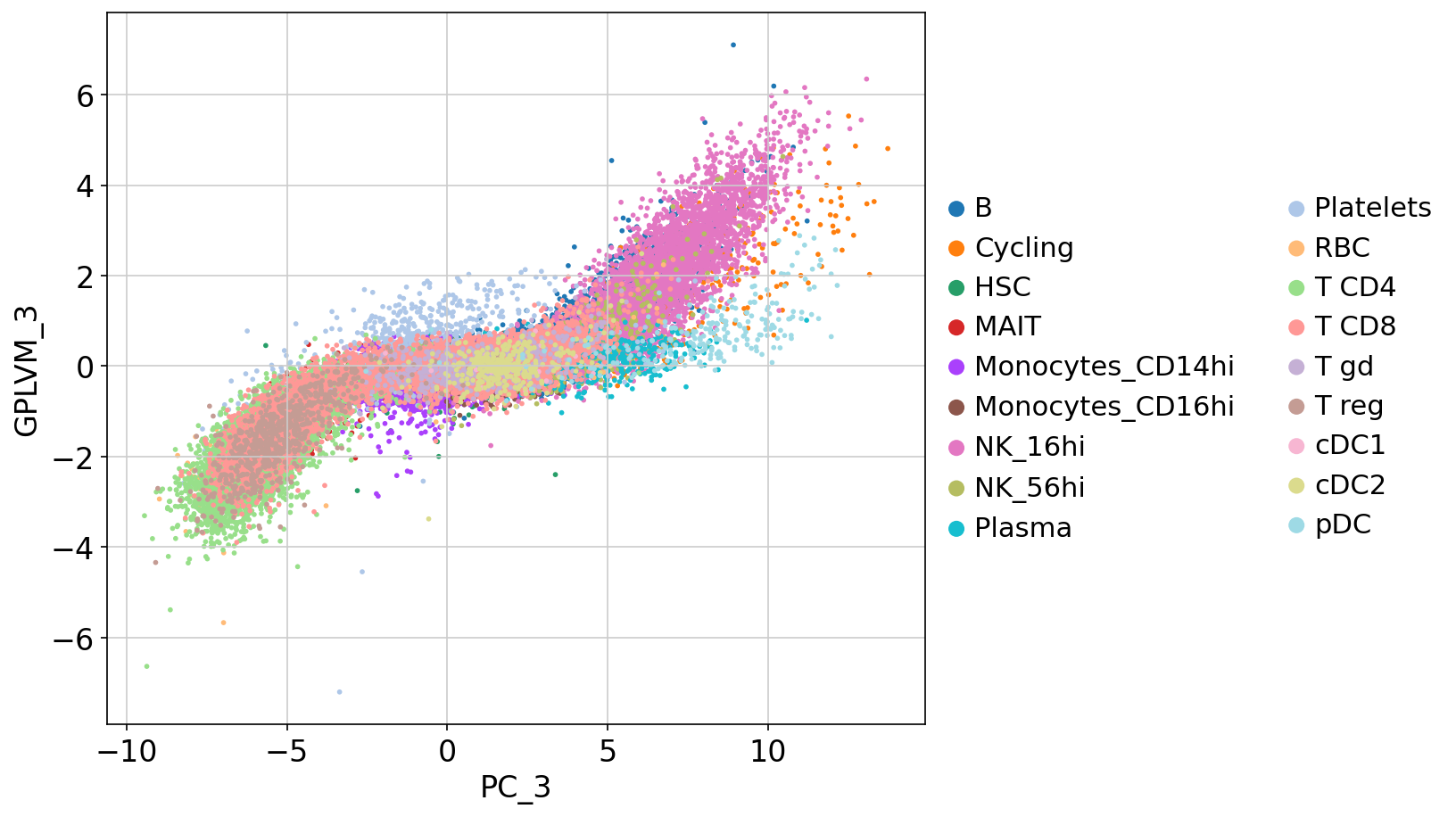}
   \includegraphics[width=0.8\textwidth]{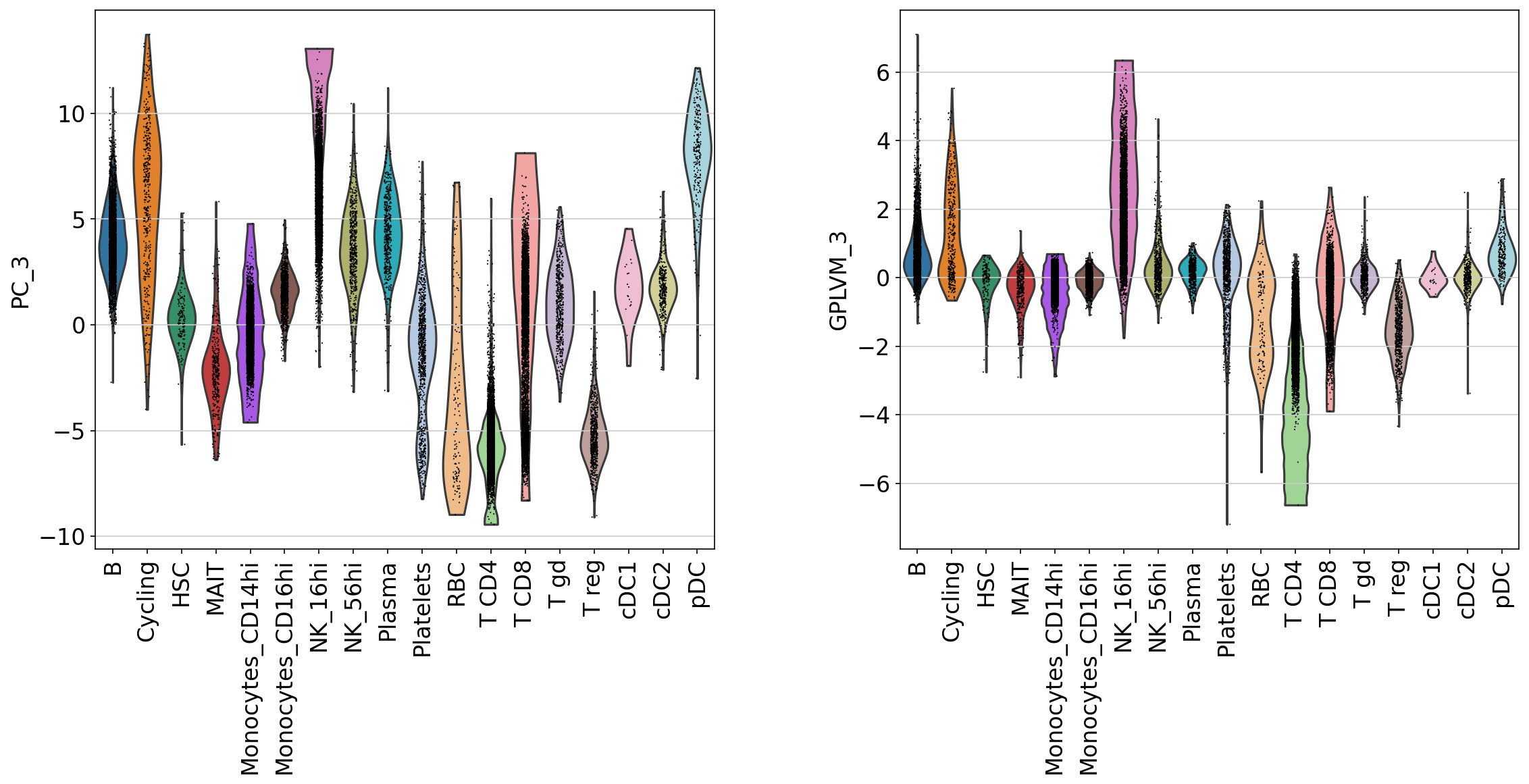}
   \includegraphics[width=0.8\textwidth]{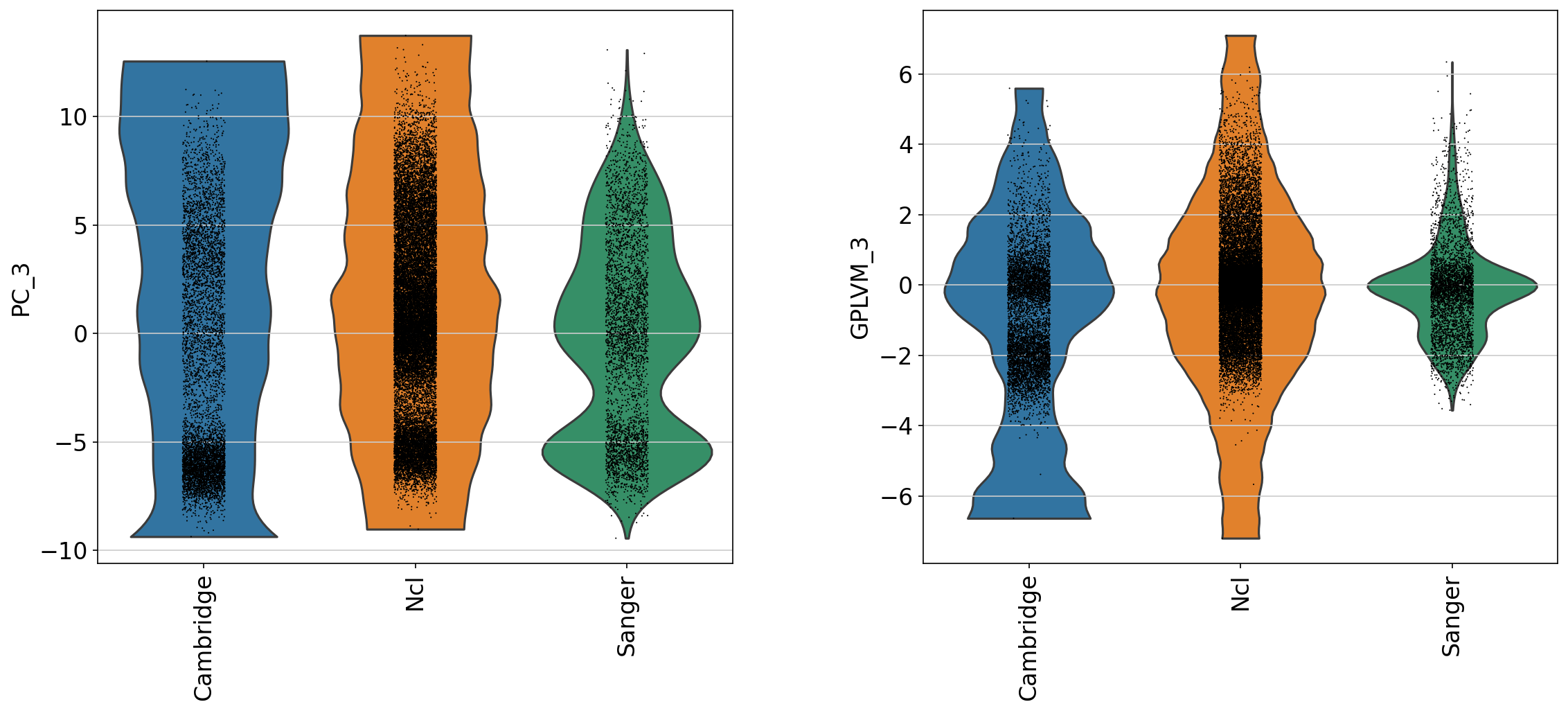}
    \caption{
      (top) Scatterplot of GPLVM latent variable 3 (GPLVM3) and the principal component used for initialization (PC3). Cells are colored by cell type. 
      (middle) Distribution of latent variable values for each cell type for PC3 (left) and GPLVM3 (right)
      (bottom) Distribution of latent variable values for each experimental batch (site of sample collection) for PC3 (left) and GPLVM3 (right)
    }
    \label{fig:pca_init}
\end{figure}

We show an example of how the GPLVM diverges from initialization in a meaningful way on the COVID dataset in Fig. ~\ref{fig:pca_init}. Here we compare one GPLVM latent variable with the principal component provided during initialization. The two are correlated and both capture cell-type specific gene expression signatures (Fig. ~\ref{fig:pca_init}B). However, in PCA the cell type specific signal is confounded with the batch effect (Fig. ~\ref{fig:pca_init}C) while with GPLVM we can intuitively use this variable to distinguish NK cells (NK\_16hi, NK\_56hi) from T cell clusters (T CD4, T CD8, T gd).

\begin{figure}
    \centering
    \includegraphics[scale=0.8]{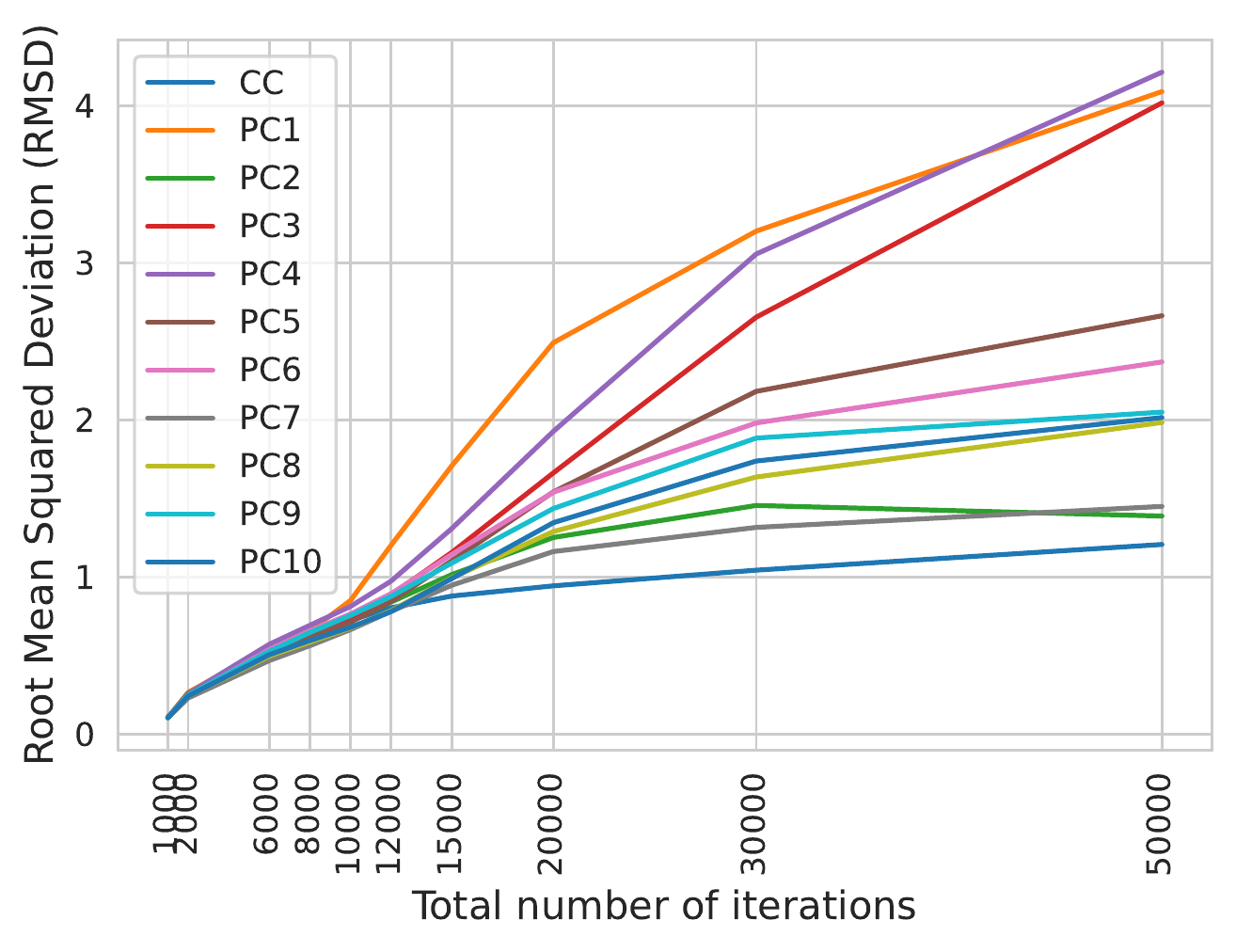}
    \caption{Latent variable deviations from initialization as the total number of epochs increases during training. CC corresponds to the cell cycle effect, while the other 10 lines correspond to the top 10 principal components of the COVID dataset.} 
    \label{fig:initial_init_test}
\end{figure}

We further note how the latent variable initialization varies as a function of the number of training iterations/epochs. Considering the first latent dimension of the GPLVM as the cell cycle (CC) effect along with 10 further dimensions initialised by the 10 PCs of the COVID dataset, Fig. \ref{fig:initial_init_test} shows how they start to deviate  significantly from the initial point as the number of iterations across the epochs increases. (The distance between an initial dimension and the optimised dimension is measured by their Root Mean Squared Deviation -- RMSD). We observe that the CC effect is the least changed overall, implying that our initialization of the average CC gene expression well defines the latent dimension that captures CC effect. PC deviations are distinctly visible,  specifically as the degree of deviation starts to spread out across a range of RMSD values.

\begin{table}[]
\centering
\caption{Training configurations for the experimental results in the manuscript. The Innate Immunity experiment used a learning rate of 0.05 for the first 3 epochs and 0.005 thereafter.}
\vspace{2mm}
\scalebox{0.9} {
\begin{tabular}{c|c|c|c|c|c|c}
         Experiment &  Learning rate & Batch size & Epochs & Latent dim $(Q)$ & Col. dim($\Phi$) & Num inducing $(M)$\\
         \hline
         Innate Immunity & 0.05 / 0.005 & 220 & 50 & 7 = 6 (rbf) + 1 (periodic) & 200 & 201 \\
         COVID w/o severity & 0.01 & 200 & 100 & 11 = 10 (rbf) + 1 (periodic) & 146 & 147 \\
         COVID w. severity & 0.005 & 300 & 50 & 12 = 10 (rbf) + 1 (periodic) + 1 (sev.) & 146 & 147  
    \end{tabular}}
    \label{tab:configs}
\end{table}

\begin{figure*}[h]
\centering
   \includegraphics[width=0.33\textwidth]{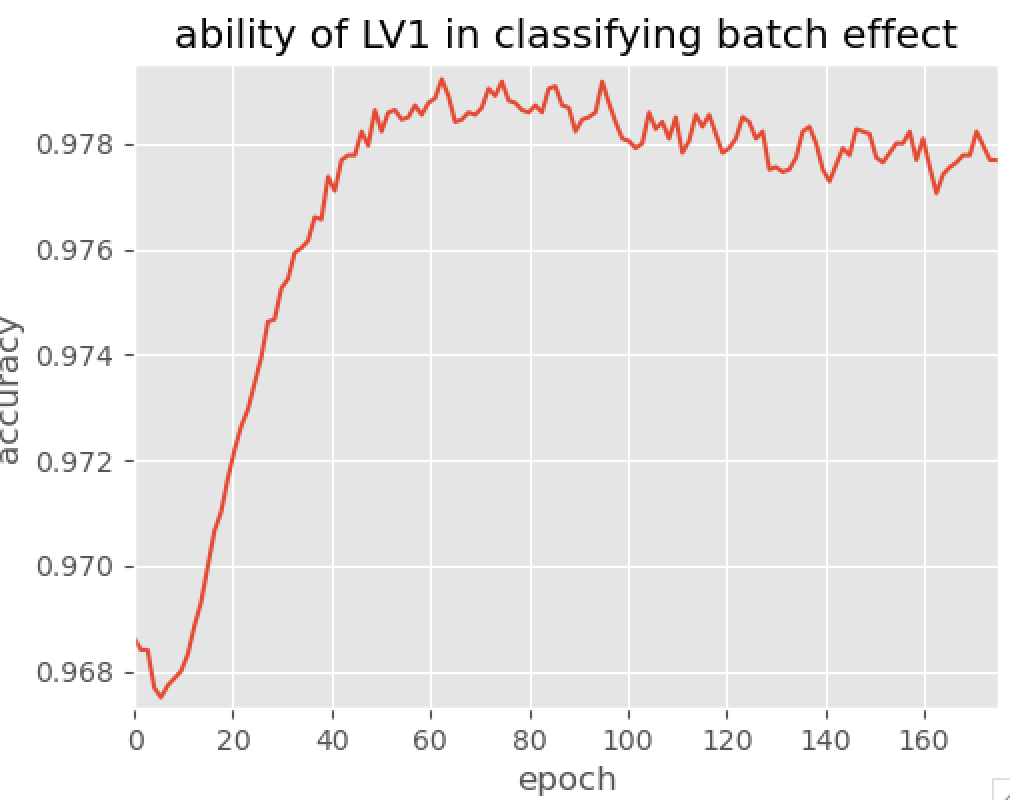}
    \caption{
        Effect of overtraining on the ability of the first latent variable (that captures batch effect) to classify batch one and two. After about a hundred epochs, the quality of the batch effect latent dimensions starts to diminish slightly.
    }
    \label{fig:overtraining}
\end{figure*}

An interesting effect observed while training the model on the Innate Immunity experiment is that overtraining can occur, albeit the effect is very subtle. \cref{fig:overtraining} shows the effect of running the model well past our stopping point (50 epochs) on the quality of the first latent variable in classifying batch effects. Quality of classification is measured as the percentage accuracy obtained using a standard classifier (a logistic regression) in being able to separate the two batches given just the first latent variable. The accuracy score drops slightly if training is continued well past 100 epochs in this experiment.

\section{Simplifying the Model to Independence across Linear and Non-linear Terms}

It's possible to simplify sparse GPs with two additive kernel components which can each be made zero by a choice of inputs (and thus, our model) so that the linear and non-linear parts of each of the $\*f_{d}$'s are independent. We drop the column index $d$ for clarity. For certain downstream tasks, we may like to interpret $\bm{f} = \*f_{1} + \*f_{2}$ as a sum of GPs, $\*f_{1}$ corresponding to the kernel $k_{per} \times k_{rbf}$ and $\*f_{2}$ corresponding to $k_{lin}$. We'd like to have the prior conditioned on inducing variables $\bm{f}|\bm{u}$ look like a sum of sparse GPs as in \citet{wilk_thesis}, $\*f | \*u = \*f_{\nonlin} | \*u_{\nonlin} + \*f_{\lin} | \*u_{\lin}$ given by, 
\begin{align}
    \label{eqn:f_decomp}
    \bm{f}|\bm{u} \sim \mathcal{N}\Big(\sum_{i \in \nonlin,\lin}K_{nm}^{(i)}K_{mm}^{{-1}^{(i)}}\bm{u}_{i}, \sum_{i \in \nonlin,\lin}K_{nn}^{(i)} - K_{nm}^{(i)}K_{mm}^{{-1}^{(i)}}K_{mn}^{(i)}\Big)
\end{align}
where $K_{nm}^{(i)}$ is the cross-covariance matrix between latent covariates $(X_{per}|X_{rbf}|\Phi)$ and $(Z_{per}|Z_{rbf}|Z_{lin})$ and with $\*u_{\lin} \perp\!\!\!\perp \*u_{\nonlin}$. The covariance matrices are given by,
\begin{align}
    K_{nm}^{\nonlin} &= k_{per}(X_{per}, Z_{per}) \times k_{rbf}(X_{rbf}, Z_{rbf}) \\
    K_{nm}^{\lin} &= k_{lin}(\Phi, Z_{lin}), \\
    K_{mm}^{\nonlin} &= k_{per}(Z_{per}, Z_{per}) \times k_{rbf}(Z_{rbf}, Z_{rbf}), \\
    K_{mm}^{\lin} &= k_{lin}(Z_{lin}, Z_{lin}).
\end{align}
This simplification can be achieved by, as an example pertaining to our kernel, partitioning the inducing points $Z$ as follows,
\begin{equation}
    Z \equiv (Z_{\nonlin}|Z_{\lin}) = \begin{bmatrix}
        Z_1 & \*0 \\
        \pm \*\infty & Z_2
    \end{bmatrix}.
\end{equation}
The zeros and infinities in this inducing inputs matrix are selected by considering inputs to a kernel that can set it to zero - for example, $k_{\text{rbf}}(., \pm\infty) = 0$ and $k_{\text{lin}}(., 0) = 0$. Due to this partitioning, the matrices of the prior,
\begin{equation}
    \*f | \*u \sim \mathcal N(K_{nm} K_{mm}^{-1} \*u, K_{nn} - K_{nm}K_{mm}^{-1}K_{mn})
\end{equation}
simplifies as follows,
\begin{align}
    K_{mm} &= \begin{bmatrix} k_{\nonlin}(Z_1, Z_1) + k_{\lin}(\*0, \*0) & k_{\nonlin}(Z_1, \pm\*\infty) + k_{\lin}(\*0, Z_2) \\ k_{\nonlin}(\pm\*\infty, Z_1) + k_{\lin}(Z_2, \*0) & k_{\nonlin}(\pm\*\infty, \pm\*\infty) + k_{\lin}(Z_2, Z_2) \end{bmatrix} = \begin{bmatrix} k_{\nonlin}(Z_1, Z_1) & \*0 \\ \*0 & k_{\lin}(Z_2, Z_2) \end{bmatrix}, \\
    K_{nm} &= \begin{bmatrix} k_{\nonlin}(X_{\nonlin}, Z_1) + k_{\lin}(\Phi, \*0) & k_{\nonlin}(X_{\nonlin}, \pm\*\infty) + k_{\lin}(\Phi, Z_2) \end{bmatrix} = \begin{bmatrix} k_{\nonlin}(X_{\nonlin}, Z_1) & k_{\lin}(\Phi, Z_2) \end{bmatrix}.
\end{align}
Cleaning up the notation,
\begin{align}
    K_{mm} = \begin{bmatrix} K^{\nonlin}_{mm} & \*0 \\ \*0 & K^{\lin}_{mm} \end{bmatrix}, 
    K_{nm} = \begin{bmatrix} K^{\nonlin}_{nm} & K^{\lin}_{nm} \end{bmatrix}.
\end{align}
By using block matrix multiplication rules, the prior simplifies to \cref{eqn:f_decomp}.
\begin{align}
    \mathbb E(\*f|\*u) &= K_{nm} K_{mm}^{-1} \*u = \begin{bmatrix} K^{\nonlin}_{nm} & K^{\lin}_{nm} \end{bmatrix} \begin{bmatrix} K^{\nonlin}_{mm} & \*0 \\ \*0 & K^{\lin}_{mm} \end{bmatrix}^{-1} \begin{bmatrix} \*u_{\nonlin} \\ \*u_{\lin} \end{bmatrix} \\
    &= K^{\nonlin}_{nm} {K^{\nonlin}}^{-1}_{mm} \*u_{\nonlin} + K^{\lin}_{nm} {K^{\lin}}^{-1}_{mm} \*u_{\lin}, \\
    \mathbb C(\*f|\*u) &= K_{nn} + K_{nm} K_{mm}^{-1} K_{mn} \\
    &= K^{\lin}_{nn} + K^{\nonlin}_{nn} + \begin{bmatrix} K^{\nonlin}_{nm} & K^{\lin}_{nm} \end{bmatrix} \begin{bmatrix} K^{\nonlin}_{mm} & \*0 \\ \*0 & K^{\lin}_{mm} \end{bmatrix}^{-1} \begin{bmatrix} K^{\nonlin}_{nm} \\ K^{\lin}_{nm} \end{bmatrix} \\
    &= (K^{\nonlin}_{nn} + K^{\nonlin}_{nm} {K^{\nonlin}}^{-1}_{mm} K^{\nonlin}_{mn}) + (K^{\lin}_{nn} + K^{\lin}_{nm} {K^{\lin}}^{-1}_{mm} K^{\lin}_{mn}).
\end{align}

Note that $k_{\text{per}}$, which is part of our kernel, doesn't matter in this simplification as if $k_{\text{rbf}}(...) = 0$, then $k_{\nonlin}(...) = k_{\text{rbf}}(...) * k_{\text{per}}(...) = 0$. Further, the number of inducing variables across $i$, $\bm{u}_{i}$'s can be different, and have shapes $(M1 \times 1)$ and $(M2 \times 1)$ without loss of generality. This formulation allows one to extract the terms in the posterior process corresponding to just the linear (random-effects) term,
%\begin{align*}
%\*Y | \*f &\sim \mathcal N(\*f, \sigma^2 \*I), \\
%\*f | \*u &\sim \mathcal N(K_{nm} K_{mm}^{-1} \*u, K_{nn} - K_{nm} K_{mm}^{-1} K_{mn}), \\
%\*u &\sim \mathcal N(\*0, K_{mm}).
%\end{align*}
%If require posterior estimates over just the linear component of $\*f$, we may restrict our model following equation 5.51 of
%$$ q(\*u_{lin}) = \mathcal N(\*u_{lin} | \*m, \*S) $$
%$$ q(\*f_{lin}) = \mathcal N(\*f_{lin} | K_{nm}K_{mm}^{-1} \*m, K_{nn} + K_{nm}K_{mm}^{-1}(\*S - %K_{mm})K_{mm}^{-1} K_{mn} ) $$
\begin{align}
q \left( \begin{bmatrix} \*u_{\not\lin} \\ \*u_{\lin} \end{bmatrix} \right) &= \mathcal N \left( \begin{bmatrix} \*u_{\not\lin} \\ \*u_{\lin} \end{bmatrix} | \begin{bmatrix} \*m_{\not\lin} \\ \*m_{\lin} \end{bmatrix}, \begin{bmatrix} S_{\not\lin} & S_c \\ S_c^T & S_{\lin} \end{bmatrix} \right), \\
q(\*f_{\lin}) &= \int p(\*f_{\lin} | \*u_{\lin}) q(\*u_{\lin}) d\*u_{\lin} = \mathcal N(\Phi Z_{\lin}^{T}(Z_{\lin}Z_{\lin}^{T})^{-1}\*m_{\lin}, \Phi( \mathbb{I} + A(S_{\lin} - Z_{\lin}Z_{\lin}^{T})A^{T}) \Phi^{T}),
\end{align}
where $A = Z_{\lin}^{T}(Z_{\lin}Z_{\lin}^{T})^{-1}$. We compute the variational posterior $q(\*f_{\lin})$ as in 
\citet{hensman2014qf}. A small jitter $\delta\mathbb{I}$ is added to make $Z_{\lin}Z_{\lin}^{T}$ invertible in practice.

\section{Supplementary Figures}
\begin{figure*}[h]
\centering
   \includegraphics[width=0.9\textwidth]{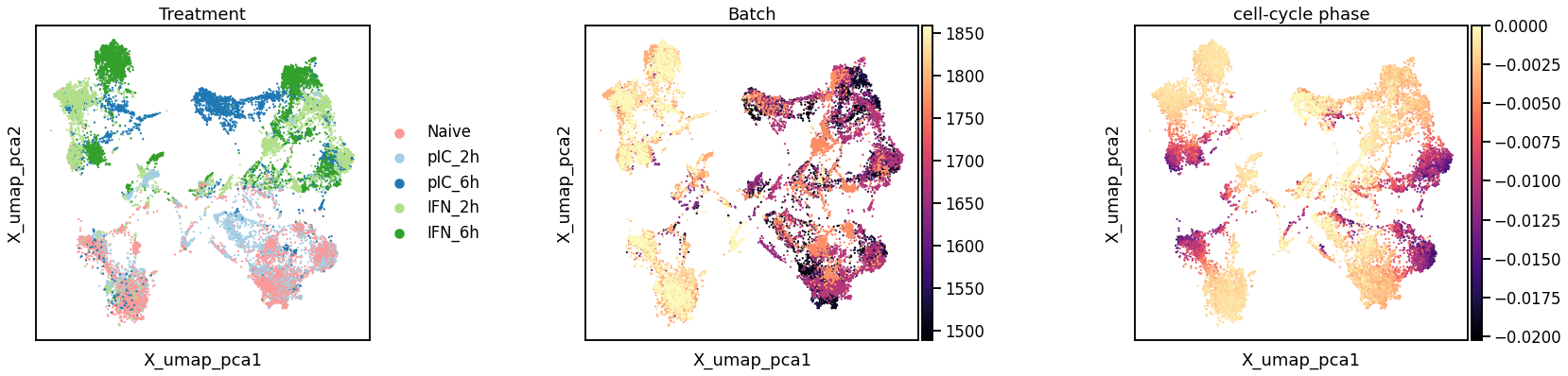}
   \includegraphics[width=0.9\textwidth]{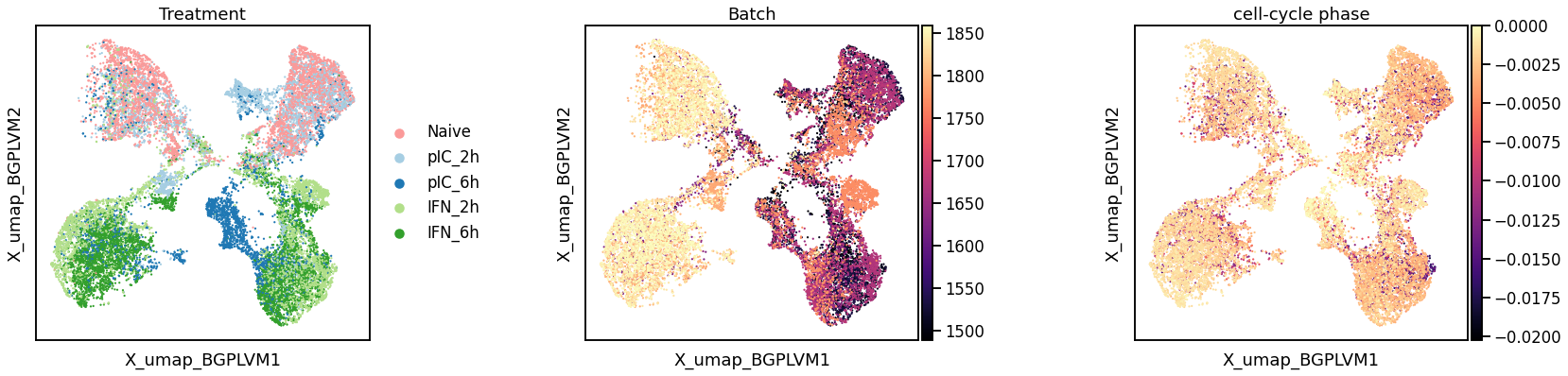}
    \caption{
        UMAP embedding of innate immunity dataset 6 dimensional latent space learnt by PCA (top) and BGPLVM (bottom). Cells are colored by treatment condition (pIC: Poly(I:C), IFN: interferon-beta) (left), experimental batch (center) and cell cycle signature (right).
    }
    \label{fig:ipscs_umaps_all}
\end{figure*}

\begin{figure}
\centering
   \includegraphics[width=\textwidth]{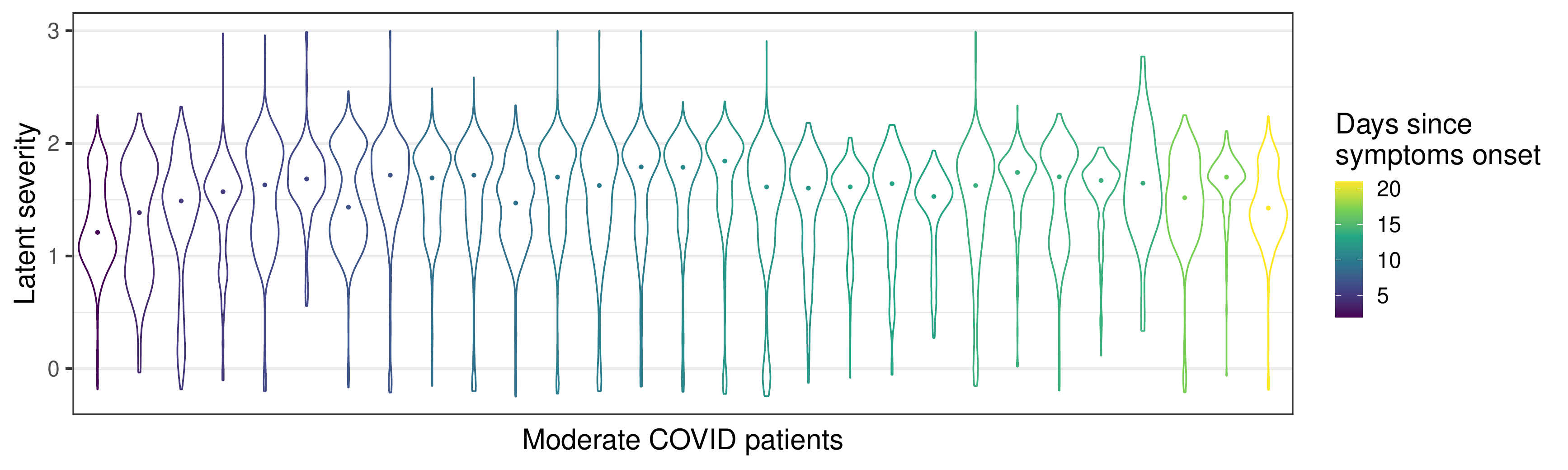}
    \caption{
        Latent COVID severity in moderate COVID patients, ordered and colored by number of days since onset of symptoms.
    }
    \label{fig:moderate_severity}
\end{figure}

% \end{appendices}

%%%%%%%%%%%%%%%%%%%%%%%%%%%%%%%%%%%%%%%%%%%%%%%%%%%%%%%%%%%%%%%%%%%%%%%%%%%%%%%
%%%%%%%%%%%%%%%%%%%%%%%%%%%%%%%%%%%%%%%%%%%%%%%%%%%%%%%%%%%%%%%%%%%%%%%%%%%%%%%

\end{document}

%% file: graphical_additive_kernel.tex
%\documentclass[tikz, border=50pt]{standalone}
%\usepackage{tikz}
%\usetikzlibrary{bayesnet}
%\usetikzlibrary{fit}
%\usepackage{bm}

%\begin{document}
\begin{tikzpicture}[scale=0.09]
\vspace{3mm}
  % Define nodes: latent, obs, det, const, factor
     \node[const, yshift=0.5cm] (Z) {$Z$};
    \node[latent, below=of Z, yshift=0.5cm] (ud) {$\bm{u}_d$};
    \node[latent, below=of ud, yshift=0.3cm] (fn) {$\bm{f}_{d}$};
    \node[latent, below=of fn, xshift=-1.4cm] (xn) {$x_{n}$};
    \node[obs, below=of fn, xshift=1.5cm] (yn) {$y_n$};
    \node[const, above=of yn] (sigma) {{$\sigma^2_{y}$}}; 
    \node[const, above=of fn, right=of ud ] (theta) {$\{\theta\}$};

    %\node[const, left=of xn, yshift=0.5cm] (mun) {$\mu_{n}$};
    %\node[const, left=of xn, yshift=-0.5cm] (sn) {$s_{n}$};
    \node[const, left=of ud, yshift=0.5cm](md){$\bm{m}_{d}$};
    \node[const, left=of ud, yshift=-0.5cm](Sd){$S_{d}$};
    \node[const, below=of xn](phi){$\phi$};

   % Connect the nodes
   \edge {xn} {fn};
   \edge {fn} {yn};
   \edge {ud} {fn};
   \edge {Z} {ud};
  % \edge{mun} {xn};
  % \edge {sn} {xn};
   \edge {md}{ud};
   \edge{Sd}{ud};
   %\edge {yn}{xn};
   %\edge {yn}{xn};
   \path[->,draw,dashed]
  (yn) edge[bend left=90] node [left] {} (xn);
  \draw[->,dashed] (phi) to node {} (xn);

   % Plates
   \plate{yf} {(yn)(xn)} {$N$} ;
    \plate{udp} {(ud)(md)(Sd)(fn)} {$D$} ;
  \edge {theta} {fn};
   \edge {sigma} {yn};
      \edge {theta} {ud};
      
\end{tikzpicture}
%\end{document}
%\end{document}

%% file: main.bbl
\begin{thebibliography}{17}
\providecommand{\natexlab}[1]{#1}
\providecommand{\url}[1]{\texttt{#1}}
\expandafter\ifx\csname urlstyle\endcsname\relax
  \providecommand{\doi}[1]{doi: #1}\else
  \providecommand{\doi}{doi: \begingroup \urlstyle{rm}\Url}\fi

\bibitem[Ahmed et~al.(2019)Ahmed, Rattray, and Boukouvalas]{ahmed2019grandprix}
Sumon Ahmed, Magnus Rattray, and Alexis Boukouvalas.
\newblock Grandprix: scaling up the {B}ayesian {GPLVM} for single-cell data.
\newblock \emph{Bioinformatics}, 35\penalty0 (1):\penalty0 47--54, 2019.

\bibitem[Buettner et~al.(2015)Buettner, Natarajan, Casale, Proserpio,
  Scialdone, Theis, Teichmann, Marioni, and
  Stegle]{buettnerComputationalAnalysisCelltocell2015}
Florian Buettner, Kedar~N. Natarajan, F.~Paolo Casale, Valentina Proserpio,
  Antonio Scialdone, Fabian~J. Theis, Sarah~A. Teichmann, John~C. Marioni, and
  Oliver Stegle.
\newblock Computational analysis of cell-to-cell heterogeneity in single-cell
  {{RNA-sequencing}} data reveals hidden subpopulations of cells.
\newblock \emph{Nature Biotechnology}, 33\penalty0 (2):\penalty0 155--160,
  February 2015.
\newblock ISSN 1546-1696.
\newblock \doi{10.1038/nbt.3102}.

\bibitem[Bui and Turner(2015)]{bui2015stochastic}
Thang~D. Bui and Richard~E. Turner.
\newblock Stochastic variational inference for {G}aussian process latent
  variable models using back constraints.
\newblock In \emph{Black Box Learning and Inference NIPS workshop}, 2015.

\bibitem[Campbell and Yau(2015)]{campbell2015bayesian}
Kieran Campbell and Christopher Yau.
\newblock Bayesian {G}aussian process latent variable models for pseudotime
  inference in single-cell rna-seq data.
\newblock \emph{bioRxiv}, page 026872, 2015.

\bibitem[Hensman et~al.(2013)Hensman, Fusi, and Lawrence]{hensman2013gaussian}
James Hensman, Nicol\'o Fusi, and Neil~D. Lawrence.
\newblock Gaussian processes for big data.
\newblock \emph{Proceedings of the Twenty-Ninth Conference on Uncertainty in
  Artificial Intelligence (UAI2013)}, 2013.

\bibitem[Hensman et~al.(2014)Hensman, Matthews, and Ghahramani]{hensman2014qf}
James Hensman, Alex Matthews, and Zoubin Ghahramani.
\newblock Scalable variational gaussian process classification, 2014.
\newblock URL \url{https://arxiv.org/abs/1411.2005}.

\bibitem[Kumasaka et~al.(2021)Kumasaka, Rostom, Huang, Polanski, Meyer, Patel,
  Boyd, Gomez, Barnett, Panousis, et~al.]{kumasaka2021mapping}
Natsuhiko Kumasaka, Raghd Rostom, Ni~Huang, Krzysztof Polanski, Kerstin Meyer,
  Sharad Patel, Rachel Boyd, Celine Gomez, Sam Barnett, Nikolaos Panousis,
  et~al.
\newblock Mapping interindividual dynamics of innate immune response at
  single-cell resolution.
\newblock \emph{bioRxiv}, 2021.

\bibitem[Lalchand et~al.(2022)Lalchand, Ravuri, and
  Lawrence]{lalchand2022generalised}
Vidhi Lalchand, Aditya Ravuri, and Neil~D. Lawrence.
\newblock Generalised {GPLVM} with {S}tochastic {V}ariational {I}nference.
\newblock In Gustau Camps-Valls, Francisco J.~R. Ruiz, and Isabel Valera,
  editors, \emph{Proceedings of The 25th International Conference on Artificial
  Intelligence and Statistics}, volume 151 of \emph{Proceedings of Machine
  Learning Research}, pages 7841--7864. PMLR, 28--30 Mar 2022.
\newblock URL \url{https://proceedings.mlr.press/v151/lalchand22a.html}.

\bibitem[Lawrence(2004)]{lawrence2004gaussian}
Neil~D. Lawrence.
\newblock Gaussian process latent variable models for visualisation of high
  dimensional data.
\newblock In \emph{Advances in neural information processing systems}, pages
  329--336, 2004.

\bibitem[Lopez et~al.(2018)Lopez, Regier, Cole, Jordan, and
  Yosef]{lopezDeepGenerativeModeling2018}
Romain Lopez, Jeffrey Regier, Michael~B. Cole, Michael~I. Jordan, and Nir
  Yosef.
\newblock Deep generative modeling for single-cell transcriptomics.
\newblock \emph{Nature Methods}, 15\penalty0 (12):\penalty0 1053--1058,
  December 2018.
\newblock ISSN 1548-7105.

\bibitem[Rasmussen and Williams(2006)]{rasmussen2004gaussian}
Carl~Edward Rasmussen and Christopher K.~I. Williams.
\newblock \emph{{Gaussian} processes in machine learning}.
\newblock Springer, 2006.

\bibitem[Stephenson et~al.(2021)Stephenson, Reynolds, Botting, {Calero-Nieto},
  Morgan, Tuong, Bach, Sungnak, Worlock, Yoshida, Kumasaka, Kania, Engelbert,
  Olabi, Spegarova, Wilson, Mende, Jardine, Gardner, Goh, Horsfall, McGrath,
  Webb, Mather, Lindeboom, Dann, Huang, Polanski, Prigmore, Gothe, Scott,
  Payne, Baker, Hanrath, {Schim van der Loeff}, Barr, {Sanchez-Gonzalez},
  Bergamaschi, Mescia, Barnes, Kilich, {de Wilton}, Saigal, Saleh, Janes,
  Smith, Gopee, Wilson, Coupland, Coxhead, Kiselev, {van Dongen}, Bacardit,
  King, Rostron, Simpson, Hambleton, Laurenti, Lyons, Meyer, Nikoli{\'c},
  Duncan, Smith, Teichmann, Clatworthy, Marioni, G{\"o}ttgens, and
  Haniffa]{stephensonSinglecellMultiomicsAnalysis2021}
Emily Stephenson, Gary Reynolds, Rachel~A. Botting, Fernando~J. {Calero-Nieto},
  Michael~D. Morgan, Zewen~Kelvin Tuong, Karsten Bach, Waradon Sungnak,
  Kaylee~B. Worlock, Masahiro Yoshida, Natsuhiko Kumasaka, Katarzyna Kania,
  Justin Engelbert, Bayanne Olabi, Jarmila~Stremenova Spegarova, Nicola~K.
  Wilson, Nicole Mende, Laura Jardine, Louis C.~S. Gardner, Issac Goh, Dave
  Horsfall, Jim McGrath, Simone Webb, Michael~W. Mather, Rik G.~H. Lindeboom,
  Emma Dann, Ni~Huang, Krzysztof Polanski, Elena Prigmore, Florian Gothe,
  Jonathan Scott, Rebecca~P. Payne, Kenneth~F. Baker, Aidan~T. Hanrath, Ina
  C.~D. {Schim van der Loeff}, Andrew~S. Barr, Amada {Sanchez-Gonzalez}, Laura
  Bergamaschi, Federica Mescia, Josephine~L. Barnes, Eliz Kilich, Angus {de
  Wilton}, Anita Saigal, Aarash Saleh, Sam~M. Janes, Claire~M. Smith, Nusayhah
  Gopee, Caroline Wilson, Paul Coupland, Jonathan~M. Coxhead, Vladimir~Yu
  Kiselev, Stijn {van Dongen}, Jaume Bacardit, Hamish~W. King, Anthony~J.
  Rostron, A.~John Simpson, Sophie Hambleton, Elisa Laurenti, Paul~A. Lyons,
  Kerstin~B. Meyer, Marko~Z. Nikoli{\'c}, Christopher J.~A. Duncan, Kenneth
  G.~C. Smith, Sarah~A. Teichmann, Menna~R. Clatworthy, John~C. Marioni,
  Berthold G{\"o}ttgens, and Muzlifah Haniffa.
\newblock Single-cell multi-omics analysis of the immune response in
  {{COVID-19}}.
\newblock \emph{Nature Medicine}, 27\penalty0 (5):\penalty0 904--916, May 2021.
\newblock ISSN 1546-170X.
\newblock \doi{10.1038/s41591-021-01329-2}.

\bibitem[Svensson et~al.(2018)Svensson, Teichmann, and
  Stegle]{svenssonSpatialDEIdentificationSpatially2018a}
Valentine Svensson, Sarah~A. Teichmann, and Oliver Stegle.
\newblock {{SpatialDE}}: Identification of spatially variable genes.
\newblock \emph{Nature Methods}, 15\penalty0 (5):\penalty0 343--346, May 2018.
\newblock ISSN 1548-7105.
\newblock \doi{10.1038/nmeth.4636}.

\bibitem[van~der Wilk(2019)]{wilk_thesis}
Mark van~der Wilk.
\newblock \emph{Sparse Gaussian Process Approximations and Applications
  (Doctoral thesis)}.
\newblock PhD thesis, University of Cambridge, 2019.

\bibitem[Velten et~al.(2022)Velten, Braunger, Argelaguet, Arnol, Wirbel,
  Bredikhin, Zeller, and Stegle]{veltenIdentifyingTemporalSpatial2022}
Britta Velten, Jana~M. Braunger, Ricard Argelaguet, Damien Arnol, Jakob Wirbel,
  Danila Bredikhin, Georg Zeller, and Oliver Stegle.
\newblock Identifying temporal and spatial patterns of variation from
  multimodal data using {{MEFISTO}}.
\newblock \emph{Nature Methods}, 19\penalty0 (2):\penalty0 179--186, February
  2022.
\newblock ISSN 1548-7105.
\newblock \doi{10.1038/s41592-021-01343-9}.

\bibitem[Verma and Engelhardt(2020)]{verma2020robust}
Archit Verma and Barbara~E Engelhardt.
\newblock A robust nonlinear low-dimensional manifold for single cell rna-seq
  data.
\newblock \emph{BMC bioinformatics}, 21\penalty0 (1):\penalty0 1--15, 2020.

\bibitem[Wolf et~al.(2018)Wolf, Angerer, and
  Theis]{wolfSCANPYLargescaleSinglecell2018}
F.~Alexander Wolf, Philipp Angerer, and Fabian~J. Theis.
\newblock {{SCANPY}}: Large-scale single-cell gene expression data analysis.
\newblock \emph{Genome Biology}, 19\penalty0 (1):\penalty0 15, February 2018.
\newblock ISSN 1474-760X.
\newblock \doi{10.1186/s13059-017-1382-0}.

\end{thebibliography}
